\documentclass[10pt,twocolumn,letterpaper]{article}

\usepackage{cvpr}              

\usepackage{graphicx}
\usepackage{amsmath}
\usepackage{amssymb}
\usepackage{booktabs}

\usepackage{times}
\usepackage{epsfig}
\usepackage{graphicx}
\usepackage{amsmath}
\usepackage{amssymb}
\usepackage{color}
\usepackage[font=footnotesize,labelfont=bf]{caption}

\usepackage{multirow}
\usepackage{adjustbox}
\usepackage{booktabs}

\newif\ifshowcomment
\showcommentfalse

\ifshowcomment
\newcommand{\qing}[1]{\textcolor{blue}{qing: #1}}
\newcommand{\felix}[1]{{\textcolor{red}{Felix: #1}}}
\newcommand{\hongkai}[1]{\textcolor{purple}{hongkai: #1}}
\newcommand{\ruijun}[1]{\textcolor{magenta}{ruijun: #1}}
\else
\newcommand{\qing}[1]{}
\newcommand{\felix}[1]{}
\newcommand{\hongkai}[1]{}
\newcommand{\ruijun}[1]{}
\fi

\def\argmax{\operatornamewithlimits{arg\,max}}

\newcommand{\rf}[1]{\textbf{\textcolor{red}{#1}}}
\newcommand{\rs}[1]{\textbf{\textcolor{green}{#1}}}
\newcommand{\rt}[1]{\textbf{\textcolor{blue}{#1}}}
\newcommand{\shrinkrow}{\renewcommand{\arraystretch}{0.856}}

\definecolor{royalblue}{RGB}{65,105,225}
\usepackage[pagebackref,breaklinks,colorlinks,citecolor=royalblue,bookmarks=false]{hyperref}

\usepackage[capitalize]{cleveref}
\crefname{section}{Sec.}{Secs.}
\Crefname{section}{Section}{Sections}
\Crefname{table}{Table}{Tables}
\crefname{table}{Tab.}{Tabs.}

\def\cvprPaperID{4405} 
\def\confName{CVPR}
\def\confYear{2022}

\begin{document}

\title{\emph{Can You Spot the Chameleon?}\\ Adversarially Camouflaging Images from Co-Salient Object Detection}

\author{Ruijun Gao\textsuperscript{1\textasteriskcentered},
Qing Guo\textsuperscript{1,6}\thanks{Ruijun Gao and Qing Guo are co-first authors and contribute equally to this work.},
~Felix Juefei-Xu\textsuperscript{2},
Hongkai Yu\textsuperscript{3},
Huazhu Fu\textsuperscript{4},
Wei Feng\textsuperscript{1}\thanks{Wei Feng is the corresponding author: \href{mailto:wfeng@ieee.org}{wfeng@ieee.org}},\\
Yang Liu\textsuperscript{5,6},
Song Wang\textsuperscript{7}
\\~\\
\textsuperscript{1}College of Intelligence and Computing, Tianjin University, China,~~ 
\textsuperscript{2}Alibaba Group, USA,~~\\
\textsuperscript{3}Cleveland State University, USA,~~
\textsuperscript{4}IHPC, A*STAR, Singapore,~~
\textsuperscript{5}Zhejiang Sci-Tech University, China,~~\\
\textsuperscript{6}Nanyang Technology University, Singapore,~~
\textsuperscript{7}University of South Carolina, USA~~
}

\maketitle


\begin{abstract}
Co-salient object detection (CoSOD) has recently achieved significant progress and played a key role in retrieval-related tasks. However, it inevitably poses an entirely new safety and security issue, \ie, highly personal and sensitive content can potentially be extracting by powerful CoSOD methods.
In this paper, we address this problem from the perspective of adversarial attacks and identify a novel task: \textbf{\emph{adversarial co-saliency attack}}. Specially, given an image selected from a group of images containing some common and salient objects, we aim to generate an adversarial version that can mislead CoSOD methods to predict incorrect co-salient regions.
Note that, compared with general white-box adversarial attacks for classification, this new task faces two additional challenges: 
(1) low success rate due to the diverse appearance of images in the group; 
(2) low transferability across CoSOD methods due to the considerable difference between CoSOD pipelines.
To address these challenges, we propose the very first black-box joint adversarial exposure and noise attack (Jadena), where we jointly and locally tune the exposure and additive perturbations of the image according to a newly designed high-feature-level contrast-sensitive loss function. 
Our method, without any information on the state-of-the-art CoSOD methods, leads to significant performance degradation on various co-saliency detection datasets and makes the co-salient objects undetectable. This can have strong practical benefits in properly securing the large number of personal photos currently shared on the Internet.
Moreover, our method is potential to be utilized as a metric for evaluating the robustness of CoSOD methods.

\end{abstract}

\section{Introduction}

\begin{figure}[t]
	\centering
	\includegraphics[width=0.9\linewidth]{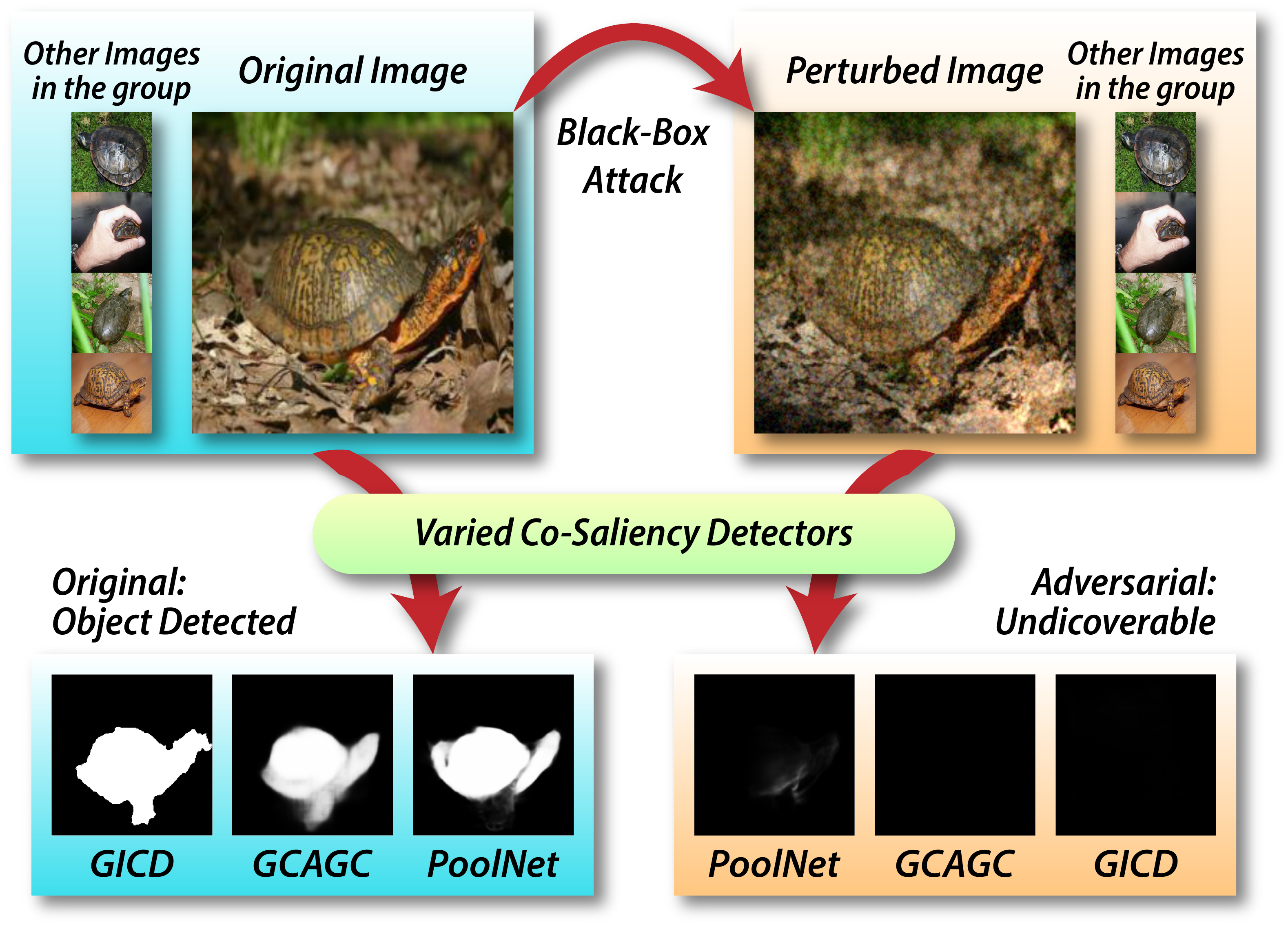}
	\caption{Overview of the novel problem, \textbf{adversarial co-saliency attack}, and our solution. We expect the perturbed image to be undiscoverable in an even dynamically growing group of images across multiple CoSOD methods (\eg, GICD \cite{zhang2020gradient}, GCAGC \cite{zhang2020adaptive}, PoolNet \cite{liu2019poolnet}), which is much more challenging and practical in real-world scenarios. Note that our attack is black-box and can be performed without references provided in the group.}
	\label{fig:overall}\vspace{-15pt}
\end{figure}

Co-saliency is typically defined as the common and salient (usually in foreground) visual stimulus found across a given group of images. Co-salient object detection (CoSOD) therefore aims at detecting and highlighting the common and salient foreground region (object) in the given group of image \cite{zhang2018review}. Different from the traditional single-image saliency detection problem, the key to solving the co-saliency problem is to discover the correspondence (based on various cues) of the common and similar salient regions across multiple images in a group. With unknown semantic categories of the co-salient objects, the co-saliency task is a rather challenging one, and a good co-saliency detection algorithm should consider both the intra-image saliency cues and the inter-image common saliency cues \cite{fu2013cluster}.

At the moment, co-saliency detection is still an emerging research topic, with many recently proposed methods to solve this challenging problem. We will discuss several of the current CoSOD algorithms, which range from low-level features to high-level semantic features, as well as deep learning-based models, in Section~\ref{sec:related}. Co-saliency detection plays a key role in many practical applications of computer vision and multimedia, including object co-segmentation \cite{zhu2016beyond}, foreground discovery in video sequences \cite{chang2015complex}, weakly supervised localization \cite{zhang2016detection}, image retrieval \cite{fu2013cluster}, multi-camera surveillance \cite{luo2015multi}, \etc. 

However, the extensive and powerful CoSOD methods inevitably pose a new safety and security problem, \ie, high-profile and personal-sensitive content may be subject to extraction and discovery by these methods. In order to prevent content-sensitive images from being discovered by co-saliency detection, in this work, we address the problem from the perspective of adversarial attacks and identify a novel task, \ie, \textbf{adversarial co-saliency attack}. Specially, given an image selected from a group containing some common and salient objects, the aim of the task is to generate an adversarial version of the image that can mislead CoSOD methods to predict incorrect co-salient regions, thus evading co-saliency detection, and making the images undiscoverable by the co-saliency detection algorithms. Our method is able to conceal personal content containing sensitive information from the CoSOD methods. Moreover, our method can potentially be used as a metric to evaluate the robustness of the CoSOD methods and can be coupled with adversarial training to improve the performance as well. 

It is worth noting that, compared with general white-box adversarial attacks for classification, this new task faces  two additional challenges:
(1) low success rate due to the diverse appearance of images in the group of images; 
(2) low transferability across CoSOD methods due to the considerable difference between CoSOD pipelines.
To overcome these challenges, we propose the very first black-box joint adversarial exposure and noise attack (Jadena), where we jointly and locally tune the exposure and additive perturbations of the image according to a newly designed high-feature-level contrast-sensitive loss function.
Our method, without needing any information on the state-of-the-art CoSOD methods, leads to significant performance degradation on various co-saliency detection datasets and makes the co-salient objects undetectable, as shown in Fig.~\ref{fig:overall}. This has strong practical value nowadays where large-scale personal multimedia contents are shared on the public domain Internet and should be properly and securely preserved and protected from malicious extraction. We have released our code in this GitHub repo:
\url{https://github.com/tsingqguo/jadena}.

\section{Related work}\label{sec:related}

{\flushleft\bf Co-saliency detection.}
As an important subarea of saliency detection \cite{achanta2009frequency,Cheng2015PAMI,LiYu16,wang2016saliency,SOD_survey}, the goal of co-saliency detection problem is to detect common salient objects in multiple images  \cite{zhang2020gradient,zhang2020adaptive,zha2020robust, jiang2019unified,li2019detecting,CoSOD3k_journal}. Co-saliency detection considers the saliency cues within images and also the common cues among images \cite{fu2013cluster,ge2016co}. Understanding correspondences of co-salient objects among multiple images is the key to co-saliency detection, which can be addressed by  optimization-based methods \cite{cao2014self,li2014repairing}, machine learning-based models  \cite{cheng2014salientshape,zhang2015self}, and deep neural networks \cite{zhang2020gradient,zhang2020adaptive,zha2020robust}. In addition, data from other modalities, such as depth images, can be fused to improve the co-saliency detection on color-based images \cite{cong2018review}. Co-saliency detection has wide applications in computer vision, including co-segmentation  \cite{jerripothula2018quality} and co-localization \cite{tang2014co}, which segments and localizes similar objects in a group of images. 

{\flushleft\bf Adversarial attack.}
With the booming of deep learning, the model safety has attracted more and more attention. Therefore many works have been proposed for  adversarial attacks against deep neural networks. Several classical methods, such as the fast gradient signed method (FGSM) \cite{goodfellow2014explaining,guo2020spark}, momentum iterative fast gradient sign method (MI-FGSM) \cite{dong2018boosting}, basic iterative method (BIM) \cite{kurakin2016adversarial}, C\&W method \cite{carlini2017towards}, show promising results. Most current attacks learn additive noise under a white-box setting, in which the model parameters are given. For example, learnable noise will be added to a clean image as a perturbation to decrease the task performance. Some other studies have also try the adversarial natural corruptions including motion blur \cite{guo2020abba,guo2021learning}, vignetting \cite{tian2021ava}, rain streaks \cite{zhai2020raining}, face morphing \cite{wang2020amora}, and haze \cite{gao2021advhaze}, and the watermark \cite{jia2020adv}. The proposed method is different from these methods because it is a non-additive noise attack based on a black-box setup.

{\flushleft\bf Adversarial attack for saliency detection.}
There is limited research on adversarial attacks and defenses for saliency detection. Tran \etal \cite{tran2020sad} have shown that adding adversarial attacks, \eg, using FGSM \cite{goodfellow2014explaining}, can result in a large performance drop for saliency detection, so they proposed a method to clean data affected by an adversarial attack. Different from attacks in image space, Che \etal \cite{che2019adversarial} have proposed a sparser adversarial perturbation in feature space against deep saliency models, which needs some model information. To the best of our knowledge, this paper is the first work on the adversarial attacks for co-saliency detection.

\section{Methodology}\label{sec:method}

\subsection{Problem formulation}

{\flushleft\bf Co-salient object detection.} Consider a dynamic group of images $\mathcal{I}=\{\mathbf{I}_i\}$ containing $|\mathcal{I}|$ images that have common and salient objects. 
`Dynamic' means the group of images can be changed or extended, since personal images are updated every day in real-world applications.
A CoSOD method denoted as $\mathrm{D}(\cdot)$ takes the group of images as input and outputs the common and salient regions in all images
%
\begin{align}\label{eq:cosod}
\mathcal{S}=\{\mathbf{S}_i\}_{i=1}^{|\mathcal{I}|}=\mathrm{D}(\mathcal{I}),
\end{align}
%
where $\mathbf{S}_i=\mathrm{D}(\mathcal{I})[i]$ is a binary map of the $i$-th image to represent its common and salient regions with `1' and others with `0'. 
Currently, most CoSOD methods have different pipelines and can achieve effective co-saliency detection.

{\flushleft\bf Task definition.} In this paper, we focus on an entirely new problem: given an image $\mathbf{I}_k$ from the dynamic group of images $\mathcal{I}$ with $k\in [1,\ldots,|\mathcal{I}|]$, we aim to make it undiscoverable to CoSOD methods.
To solve this problem, we formulate it as an adversarial attack, applying adversarial perturbations to image $\mathbf{I}_k$ to fool a CoSOD method $\mathrm{D}(\cdot)$ into predicting an incorrect salient map, \ie, $\mathbf{S}_k$.
Specifically, we can add perturbations to the original image $\mathbf{I}_k$ by $\mathbf{I}_k^\mathrm{a}=P_\theta(\mathbf{I}_k)$, where $\mathbf{I}_k^\mathrm{a}$ is the adversarial example aiming to fool the CoSOD method and $P_\theta(\mathbf{I}_k)$ is an image transformation (\eg, $P_\theta(\mathbf{I}_k)=\mathbf{I}_k+\theta$ for the popular additive-perturbation-based attack).
The key to the adversarial attack is how to calculate the parameter $\theta$.

\subsection{Adversarial noise attack and challenges} \label{sub:adv_noise}
We first consider a straightforward way to address the task with the popular additive-perturbation-based attack.  
To be specific, we define 
%
\begin{align}\label{eq:add}
\mathbf{I}_k^\mathrm{a}=P_\theta(\mathbf{I}_k)=\mathbf{I}_k+\theta,
\end{align}
%
where $\theta$ is a pixel-wise perturbation of the same size as $\mathbf{I}_k$. Then, we can define the following objective function for calculating $\theta$:
%
\begin{align}\label{eq:adv_noise}
\argmax_{\theta} &~  J(\mathrm{D}(\{\mathbf{I}_k+\theta\}\cup\{\mathbf{I}_i|i\neq k\})[k], \mathbf{S}_k),\nonumber \\ 
\, & \text{subject to~~} \, \|\theta\|_\mathrm{p}\leq\epsilon,
\end{align}
where $\mathbf{S}_k$ is the CoSOD result of $\mathbf{I}_k$ based on the original group of images and $J(\cdot)$ is a loss function measuring the distance between the predictive results and $\mathbf{S}_k$. The constraint function, \ie, $\|\theta\|_\mathrm{p}\leq\epsilon$, is used to control the distortion degree determined by $\theta$.

We can optimize the above function with the standard optimizations for additive-perturbation-based attacks, \ie, using one-step signed gradient-based, iterative-based, and momentum-based optimization methods.
However, we find such a general adversarial noise attack cannot address the new task very well since it faces two \textit{challenges}:
(1) It is difficult for the white-box adversarial noise attack to reach a high success rate since the dynamic group of images $\mathcal{I}$ makes the optimized $\theta$ ineffective when new images are added. 
(2) Because there are several diverse pipelines for CoSOD, the optimized $\theta$ has low transferability. That is, the optimized $\theta$ for one CoSOD method fails to fool another.
%

\subsection{Joint adversarial exposure and noise attack}

In this section, we introduce a new black-box attack method that can address the two challenges faced by white-box adversarial noise attacks. 
In general, the saliency of an object is related to its contrast to other regions of the input image, while the contrast is usually affected by the image exposure. 
To achieve an effective attack, rather than only considering the noise, as in Sec.~\ref{sub:adv_noise}, we propose to jointly consider exposure and noise factors during the attack.

\begin{figure}[t]
	\centering
	\includegraphics[width=0.95\linewidth]{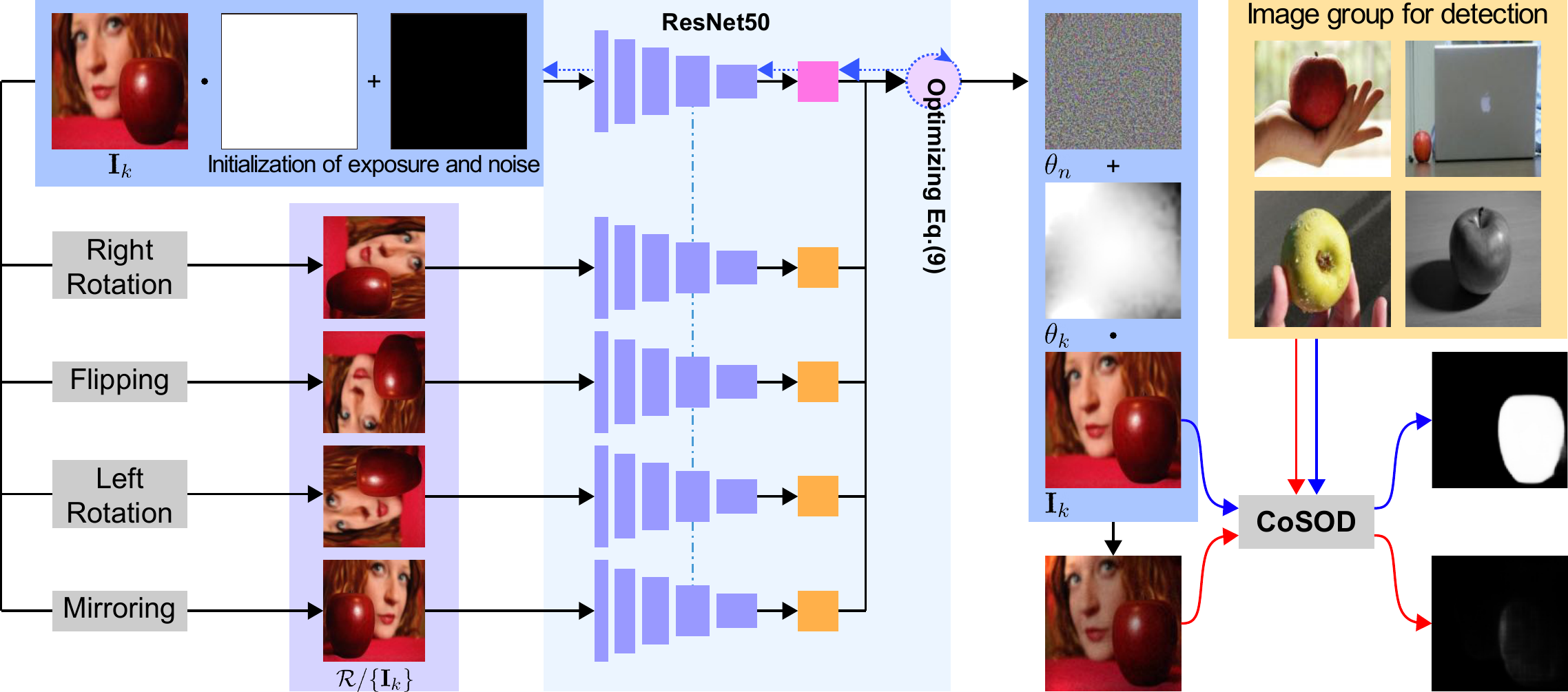}
	\vspace{-8pt}
	\caption{Pipeline of the \textit{joint adversarial exposure and noise attack}. The clean image is augmented to generate references and the gradient back-propagates along the blue dashed lines.}
	\label{fig:pipeline}\vspace{-15pt}
\end{figure}

{\flushleft\bf Formulation.} We add the joint exposure and noise based perturbations to the image $\mathbf{I}_k$ by
%
\begin{align}\label{eq:adv_attack}
\mathbf{I}^{\mathrm{a}}_k = P_\theta(\mathbf{I}_k)=\theta_e\mathbf{I}_k+\theta_n,
\end{align}
%
where $\theta=[\theta_e,\theta_n]$ and $\theta_e$ and $\theta_n$ denote the pixel-wise exposure and noise matrices, respectively.
Note that the above exposure model has been widely used in various image enhancement methods \cite{fu2016cvpr,zhang2020tmm}. However, achieving an effective co-saliency attack is still challenging: (1) The exposure is smooth across the whole image domain and cannot be tuned pixel-wise, as this will result in a speckle-like appearance. (2) We still do not know how to optimize $\theta$ to achieve a black-box attack across different CoSOD methods.

For the first challenge, we construct an objective function to encourage the naturalness of the exposure. For the second, instead of optimizing $\theta$ based on a CoSOD method, we explore a full black-box approach, where we define the objective functions on the high-level features extracted from a pretrained classification model denoted as $\phi(\cdot)$. 

{\flushleft\bf Objective function for the naturalness of the adversarial exposure.} We emphasize that exposure in the real world is usually smooth (\ie, neighboring pixels tend to have very similar exposure) across the whole image. In contrast, the adversarial attacks require the exposure to be locally different for a high success rate. 
Besides, we want the exposure to modify the original input as little as possible.
To meet these requirements, we propose to represent the exposure with the locally-variant multivariate polynomial model in the logarithm domain:
%
\begin{align}\label{eq:obj_exposure_poly}
\log(\theta_{e,p})=\sum_{d=0}^{D}\sum_{l=0}^{D-d}a_{d,l}(x_p+u_p)^d(y_p+v_p)^l,
\end{align}
%
where $x_p$ and $y_p$ are the coordinates of the $p$-th pixel and $u_p$ and $v_p$ denote the variation offsets of the $p$-th pixel.
$\{a_{d,l}\}$ denotes the parameters of the polynomial model and is concatenated as $\mathbf{a}$ with $D$ being the degree.
Note that a polynomial model with fewer parameters $\{a_{d,l}\}$ leads to smoother exposure.
With Eq.~\eqref{eq:obj_exposure_poly}, we can represent the exposure as a function of $\mathbf{a}$ and the offset map $\mathbf{U}=\{(u_p,v_p)\}$, \ie, $\theta_e(\mathbf{a},\mathbf{U})$.
More specifically, to maintain the smoothness of exposure and make it capable of an adversarial attack, we define the following objective function:  
\begin{align}\label{eq:obj_smooth}
J_{\text{smooth}}(\mathbf{a},\mathbf{U}) = -\lambda_b\|\log(\theta_e(\mathbf{a},\mathbf{U}))\|_2^2-\lambda_s\|\nabla\mathbf{U}\|_2^2,
\end{align}
%
where the first term encourages the model to not perturb the original $\mathbf{I}_k$ and the second term is a total variation regularization, which encourages smooth variation of the exposure.

{\flushleft\bf Objective function against single saliency.} We argue that the image $\mathbf{I}_k$ should be perturbed by the exposure and noise to make its high-level features consistent across all image regions. As a result, the salient objects will not be invisible at a high level. To this end, we define the objective function against single saliency: 
%
\begin{align}\label{eq:obj_consist}
J_{\text{cons}}(\mathbf{a}, \mathbf{U}, \theta_n) = -\text{avg}(\{\text{std}(\phi_j(P_\theta(\mathbf{I}_k)))|j\in\mathcal{L}\}) 
\end{align}
%
where $\phi_j(\cdot)$ is the $j$-th layer feature of the CNN $\phi(\cdot)$ and $\mathcal{L}$ denotes the index set of employed layers. The function $\text{std}(\cdot)$ calculates the standard variation of each feature channel and outputs the average across all channels. The function $\text{avg}(\cdot)$ computes the average of all $\text{std}(\phi_j(P_\theta(\mathbf{I}_k)))$ across all required layers.

{\flushleft\bf Objective function against co-saliency.} Eq.~\eqref{eq:obj_consist} is used to make the object less salient in a single image and may be less effective for co-saliency detection since the non-salient object could be salient again in a group of images. We thus extend Eq.~\eqref{eq:obj_consist} to the feature level of a group of images, \ie, $\mathcal{R}$ with $P_\theta(\mathbf{I}_k)\in\mathcal{R}$, and have
%
\begin{align}\label{eq:obj_coconsist}
J_{\text{co-cons}}(\mathbf{a}, \mathbf{U}, \theta_n) &= -\text{avg}(\{\text{std}(\Phi_j(\mathcal{R}))|j\in\mathcal{L}\}),  \nonumber \\
\textrm{with~~~~}\Phi_j(\mathcal{R}) &=\text{spl}(\{\phi_j(\mathbf{I}_i)|\mathbf{I}_i\in\mathcal{R}\}).
\end{align}
%
Note that we splice the $j$-th layer features of all images in $\mathcal{R}$ with the function $\text{spl}(\cdot)$, in a channel-wise way. For example, if the function $\phi_j(\mathbf{I}_i)$ outputs a tensor with size $11\times 11\times 256$ for $|\mathcal{R}|=4$, we get $\Phi_j(\mathcal{R})$ of size $44\times 11\times 256$. Intuitively, compared with Eq.~\eqref{eq:obj_consist}, Eq.~\eqref{eq:obj_coconsist} makes the salient regions in $\mathbf{I}_k$ become less salient across the group $\mathcal{R}$. We consider two ways of constructing $\mathcal{R}$. One uses the transformation of $\mathbf{I}_k$, \eg, rotation and up-down flip; the other employs other images. We will show that both can achieve effective co-saliency attack.
%

{\flushleft\bf Optimization.} With the above objective functions, we can calculate the adversarial exposure (\ie, $\theta_e(\mathbf{a},\mathbf{U})$) and noise (\ie,  $\theta_n$) by optimizing
%
\begin{align}\label{eq:adv_exp_noise}
\argmax_{\mathbf{a}, \mathbf{U}, \theta_n} &~  J_{\text{co-cons}/\text{cons}}(\mathbf{a}, \mathbf{U}, \theta_n)+J_{\text{smooth}}(\mathbf{a},\mathbf{U}) ,\nonumber \\ 
\, & \text{subject to~~} \, \|\theta_n\|_\mathrm{p}\leq\epsilon.
\end{align}
%
Compared with the objective function used by the adversarial noise attack, \ie, Eq.~\eqref{eq:adv_noise}, the new objective function has the following advantages: (1) It does not rely on CoSOD methods or the networks they use. This means that our method is a full black-box attack and could be very useful in practical applications. (2) Our method is unsupervised since it does not use the saliency map, \ie, $\mathbf{S}_k$ in Eq.~\eqref{eq:adv_noise}, which also makes it more useful in the real world. (3) We use both exposure and noise for a more effective attack. We employ the sign gradient descent algorithm to optimize Eq.~\eqref{eq:adv_exp_noise}.

{\flushleft\bf Algorithm.} As shown in Fig.~\ref{fig:pipeline}, our method takes a clean image as input $\mathbf{I}_k$. In addition to fetching references from a group of images, we design another approach in which $\mathbf{I}_k$ is augmented by flipping, mirroring, left rotation, and right rotation to generate four reference images. We collect the original image and references together to build $\mathcal{R}$.
Then, we initialize the coefficients $\mathbf{a}$ of the exposure polynomial model, offset map $\mathbf{U}$, and pixel-wise noise $\theta_n$, all with zero values. We follow Eq.~\eqref{eq:adv_attack} and Eq.~\eqref{eq:obj_exposure_poly} to perturb the clean image. Note that only the original image $\mathbf{I}_k$ is perturbed. For initialization, the perturbation is an identity operation.
We employ a sign gradient descent algorithm to optimize Eq.~\eqref{eq:adv_exp_noise} and MI-FGSM is adopted in this paper.
Finally, the optimized $\mathbf{a}$, $\mathbf{U}$, and $\theta_n$ are applied to the original clean image to generate a perturbed image as the output, which may fool CoSOD methods.

\section{Experiments}\label{sec:exp}


\begin{table*}[tb]
    \shrinkrow
    \centering
    \begin{adjustbox}{width=\linewidth,center}
    \begin{tabular}{cl|cccc|cccc|cccc|cccc}
    \toprule
      \multicolumn{2}{c|}{}
    & \multicolumn{4}{c|}{GICD}
    & \multicolumn{4}{c|}{GCAGC}
    & \multicolumn{4}{c|}{CBCD}
    & \multicolumn{4}{c}{PoolNet} \\
    &
    & S $\uparrow$ & AP $\downarrow$ & $F_\beta$ $\downarrow$ & MAE $\uparrow$
    & S $\uparrow$ & AP $\downarrow$ & $F_\beta$ $\downarrow$ & MAE $\uparrow$
    & S $\uparrow$ & AP $\downarrow$ & $F_\beta$ $\downarrow$ & MAE $\uparrow$
    & S $\uparrow$ & AP $\downarrow$ & $F_\beta$ $\downarrow$ & MAE $\uparrow$ \\ \midrule\midrule
\multirow{6}{*}{\rotatebox[origin=c]{90}{Cosal2015}}&
Original        &
0.2352        & 0.8595        & 0.7800        & 0.0838        &
0.1881        & 0.8960        & 0.8275        & 0.0814        &
0.9504        & 0.6046        & 0.1530        & 0.2287        &
0.2625        & 0.8449        & 0.7626        & 0.1001        \\
&
Jadena$_\mathrm{\emph{single}}$&
0.6864        & 0.7044        & 0.3831        & 0.1753        &
0.4868        & 0.7682        & 0.6162        & 0.1529        &
0.9663        & 0.5490        & 0.1199        & 0.2354        &
0.5538        & 0.6985        & 0.5152        & 0.1431        \\
&
Jadena$_\mathrm{\emph{group}}$&
0.6849        & 0.7069        & 0.3838        & \rf{0.1777}   &
\rf{0.5221}   & 0.7453        & \rf{0.5842}   & \rf{0.1619}   &
0.9826        & 0.5212        & 0.0715        & 0.2392        &
\rf{0.6060}   & 0.6671        & 0.4650        & \rf{0.1545}   \\
&
Jadena$_\mathrm{\emph{augment}}$&
\rf{0.6893}   & \rf{0.7016}   & \rf{0.3747}   & 0.1749        &
0.5146        & \rf{0.7432}   & 0.5960        & 0.1592        &
0.9816        & 0.5137        & 0.0754        & 0.2394        &
0.6010        & \rf{0.6648}   & \rf{0.4616}   & 0.1513        \\
&
Jedena$_\mathrm{\emph{augment}}$ w/o noise&
0.3226        & 0.8297        & 0.7074        & 0.1037        &
0.2010        & 0.8860        & 0.8182        & 0.0870        &
\rf{0.9921}   & \rf{0.4853}   & \rf{0.0434}   & \rf{0.2422}   &
0.3677        & 0.8032        & 0.6824        & 0.1145        \\
&
Jedena$_\mathrm{\emph{augment}}$ w/o exposure&
0.5261        & 0.7536        & 0.5483        & 0.1437        &
0.3821        & 0.8185        & 0.7074        & 0.1243        &
0.9533        & 0.6028        & 0.1473        & 0.2290        &
0.4303        & 0.7680        & 0.6347        & 0.1240        \\\midrule
\multirow{6}{*}{\rotatebox[origin=c]{90}{iCoseg}}&
Original        &
0.2970        & 0.8189        & 0.7761        & 0.0851        &
0.1991        & 0.8721        & 0.8036        & 0.0788        &
0.7760        & 0.8005        & 0.4470        & 0.1744        &
0.2286        & 0.8845        & 0.7955        & 0.0709        \\
&
Jadena$_\mathrm{\emph{single}}$&
0.6952        & 0.7063        & 0.4231        & 0.1666        &
0.3142        & 0.8369        & 0.7257        & 0.1111        &
0.8383        & 0.7422        & 0.3621        & 0.1849        &
0.5381        & 0.7677        & 0.5576        & 0.1346        \\
&
Jadena$_\mathrm{\emph{group}}$&
0.6874        & 0.7070        & 0.4242        & 0.1663        &
0.3453        & 0.8167        & 0.7046        & 0.1166        &
0.9036        & 0.7073        & 0.2425        & 0.1979        &
\rf{0.5848}   & 0.7474        & \rf{0.4986}   & \rf{0.1450}   \\
&
Jadena$_\mathrm{\emph{augment}}$&
\rf{0.7092}   & \rf{0.7017}   & \rf{0.3966}   & \rf{0.1708}   &
\rf{0.3733}   & \rf{0.8130}   & \rf{0.6989}   & \rf{0.1178}   &
0.9051        & 0.7089        & 0.2324        & 0.1961        &
0.5785        & \rf{0.7351}   & 0.5002        & 0.1435        \\
&
Jedena$_\mathrm{\emph{augment}}$ w/o noise&
0.3997        & 0.7905        & 0.7171        & 0.1038        &
0.1975        & 0.8636        & 0.8015        & 0.0821        &
\rf{0.9425}   & \rf{0.6434}   & \rf{0.1361}   & \rf{0.2068}   &
0.3515        & 0.8456        & 0.6965        & 0.0957        \\
&
Jedena$_\mathrm{\emph{augment}}$ w/o exposure&
0.5490        & 0.7525        & 0.5593        & 0.1406        &
0.2457        & 0.8626        & 0.7736        & 0.0919        &
0.7916        & 0.7998        & 0.4307        & 0.1750        &
0.3670        & 0.8405        & 0.7051        & 0.1020        \\\midrule
\multirow{6}{*}{\rotatebox[origin=c]{90}{CoSOD3k}}&
Original        &
0.3555        & 0.7963        & 0.6981        & 0.0927        &
0.2995        & 0.8273        & 0.7672        & 0.0949        &
0.9717        & 0.5184        & 0.1400        & 0.2024        &
0.3667        & 0.7717        & 0.6947        & 0.1164        \\
&
Jadena$_\mathrm{\emph{single}}$&
0.7630        & 0.6592        & 0.3282        & 0.1613        &
0.6092        & 0.6878        & 0.5481        & 0.1519        &
0.9795        & 0.4692        & 0.1080        & 0.2070        &
0.6622        & 0.6065        & 0.4425        & 0.1513        \\
&
Jadena$_\mathrm{\emph{group}}$&
0.7672        & 0.6600        & 0.3158        & \rf{0.1638}   &
\rf{0.6441}   & 0.6630        & \rf{0.5125}   & \rf{0.1589}   &
0.9873        & 0.4411        & 0.0632        & 0.2095        &
0.7156        & 0.5681        & \rf{0.3823}   & 0.1579        \\
&
Jadena$_\mathrm{\emph{augment}}$&
\rf{0.7741}   & \rf{0.6544}   & \rf{0.3075}   & 0.1630        &
0.6360        & \rf{0.6601}   & 0.5153        & 0.1586        &
0.9879        & 0.4400        & 0.0741        & \rf{0.2096}   &
\rf{0.7189}   & \rf{0.5614}   & 0.3827        & \rf{0.1590}   \\
&
Jedena$_\mathrm{\emph{augment}}$ w/o noise&
0.4484        & 0.7657        & 0.6307        & 0.1075        &
0.3139        & 0.8154        & 0.7467        & 0.0993        &
\rf{0.9934}   & \rf{0.4119}   & \rf{0.0341}   & 0.2095        &
0.4756        & 0.7192        & 0.6124        & 0.1265        \\
&
Jedena$_\mathrm{\emph{augment}}$ w/o exposure&
0.6345        & 0.6959        & 0.4627        & 0.1404        &
0.5069        & 0.7361        & 0.6235        & 0.1316        &
0.9744        & 0.5158        & 0.1367        & 0.2030        &
0.5516        & 0.6779        & 0.5530        & 0.1376        \\\midrule
\multirow{6}{*}{\rotatebox[origin=c]{90}{CoCA}}&
Original        &
0.6680        & 0.5993        & 0.4939        & 0.1187        &
0.6062        & 0.5855        & 0.5970        & 0.1153        &
0.9606        & 0.3543        & 0.1738        & 0.1377        &
0.7259        & 0.4923        & 0.4707        & \rf{0.1681}   \\
&
Jadena$_\mathrm{\emph{single}}$&
0.8046        & 0.5332        & 0.3650        & \rf{0.1265}   &
0.7660        & 0.4733        & 0.4893        & 0.1315        &
0.9737        & 0.3228        & 0.1455        & \rf{0.1401}   &
0.7954        & 0.4283        & 0.3829        & 0.1320        \\
&
Jadena$_\mathrm{\emph{group}}$&
0.8340        & 0.5327        & \rf{0.3335}   & 0.1226        &
\rf{0.7915}   & \rf{0.4542}   & 0.4584        & 0.1341        &
\rf{0.9876}   & 0.2998        & 0.0974        & 0.1380        &
\rf{0.8355}   & \rf{0.3853}   & \rf{0.3221}   & 0.1271        \\
&
Jadena$_\mathrm{\emph{augment}}$&
\rf{0.8363}   & \rf{0.5284}   & 0.3365        & 0.1224        &
0.7876        & 0.4588        & \rf{0.4502}   & \rf{0.1347}   &
0.9846        & 0.2973        & 0.0995        & 0.1392        &
0.8270        & 0.3909        & 0.3399        & 0.1284        \\
&
Jedena$_\mathrm{\emph{augment}}$ w/o noise&
0.6996        & 0.5852        & 0.4664        & 0.1120        &
0.6100        & 0.5737        & 0.5963        & 0.1179        &
0.9876        & \rf{0.2833}   & \rf{0.0643}   & 0.1352        &
0.7560        & 0.4571        & 0.4394        & 0.1541        \\
&
Jedena$_\mathrm{\emph{augment}}$ w/o exposure&
0.7490        & 0.5535        & 0.4310        & 0.1250        &
0.7112        & 0.5244        & 0.5536        & 0.1222        &
0.9691        & 0.3526        & 0.1672        & 0.1385        &
0.7598        & 0.4588        & 0.4428        & 0.1405        \\\bottomrule
    \end{tabular}
    \end{adjustbox}
    \vspace{-8pt}
    \caption{Attack performance. ``Original'' means that we predict the co-saliency map of clean images using each CoSOD method. Jadena$_{\left\{\mathrm{\emph{single}},\,\mathrm{\emph{augment}},\,\mathrm{\emph{group}}\right\}}$ indicates which variant of Jadena is used, and ``w/o noise'' or ``w/o exposure'' means that the attack is performed without additive noise or exposure tuning. We highlight the top results of each CoSOD method and each dataset in \rf{red}.}
    \label{tab:atk_perf}\vspace{-8pt}
\end{table*}

\begin{table*}[tb]
    \shrinkrow
    \centering
    \begin{adjustbox}{width=\linewidth,center}
    \begin{tabular}{l|cccc|cccc|cccc|cccc}
    \toprule
    & \multicolumn{4}{c|}{GICD}
    & \multicolumn{4}{c|}{GCAGC}
    & \multicolumn{4}{c|}{CBCD}
    & \multicolumn{4}{c}{PoolNet} \\
    & S $\uparrow$ & AP $\downarrow$ & $F_\beta$ $\downarrow$ & MAE $\uparrow$
    & S $\uparrow$ & AP $\downarrow$ & $F_\beta$ $\downarrow$ & MAE $\uparrow$
    & S $\uparrow$ & AP $\downarrow$ & $F_\beta$ $\downarrow$ & MAE $\uparrow$
    & S $\uparrow$ & AP $\downarrow$ & $F_\beta$ $\downarrow$ & MAE $\uparrow$ \\ \midrule\midrule
Original        &
0.2352         & 0.8595         & 0.7800         & 0.0838         &
0.1881         & 0.8960         & 0.8275         & 0.0814         &
0.9504         & 0.6046         & 0.1530         & 0.2287         &
0.2625         & 0.8449         & 0.7626         & 0.1001         \\\midrule\midrule
Noise$_{8}$     &
0.2238         & 0.8602         & 0.7863         & 0.0841         &
0.2074         & 0.8893         & 0.8179         & 0.0868         &
0.9464         & 0.6057         & 0.1535         & 0.2283         &
0.2655         & 0.8432         & 0.7628         & 0.1011         \\
Noise$_{16}$    &
0.2367         & 0.8525         & 0.7734         & 0.0886         &
0.2323         & 0.8779         & 0.8009         & 0.0950         &
0.9504         & 0.6054         & 0.1476         & 0.2289         &
0.2844         & 0.8358         & 0.7521         & 0.1048         \\
Noise$_{32}$    &
0.3012         & 0.8216         & 0.7310         & 0.1048         &
0.2988         & 0.8395         & 0.7519         & 0.1200         &
0.9598         & 0.5953         & 0.1376         & 0.2313         &
0.3429         & 0.8144         & 0.7078         & 0.1145         \\\midrule\midrule
FGSM$_\mathrm{\emph{BCE}}$&
0.4422        *& 0.7644        *& 0.6366        *& 0.1351        *&
0.3355         & 0.8309         & 0.7315         & 0.1295         &
0.9543         & 0.5961         & 0.1503         & 0.2302         &
0.3454         & 0.8089         & 0.7043         & 0.1175         \\
FGSM$_{L_1}$    &
0.4561        *& 0.7655        *& 0.6098        *& 0.1410        *&
0.3300         & 0.8342         & 0.7384         & 0.1294         &
0.9573         & 0.5982         & 0.1458         & 0.2302         &
0.3524         & 0.8077         & 0.6998         & 0.1168         \\\midrule
I-FGSM$_\mathrm{\emph{BCE}}$&
\rf{0.7285}   *& \rf{0.6510}   *& 0.3970        *& \rf{0.3337}   *&
0.2769         & 0.8455         & 0.7644         & 0.1146         &
0.9533         & 0.6042         & 0.1491         & 0.2286         &
0.2754         & 0.8339         & 0.7503         & 0.1057         \\
I-FGSM$_{L_1}$  &
0.6397        *& \rt{0.6561}   *& 0.5134        *& \rt{0.2700}   *&
0.2615         & 0.8643         & 0.7801         & 0.1063         &
0.9454         & 0.6059         & 0.1533         & 0.2280         &
0.2715         & 0.8364         & 0.7533         & 0.1052         \\\midrule
MI-FGSM$_\mathrm{\emph{BCE}}$ &
0.6700        *& 0.6697        *& 0.4480        *& \rs{0.2868}   *&
0.3975         & 0.7909         & 0.6842         & \rt{0.1532}    &
0.9514         & 0.5995         & 0.1468         & 0.2293         &
0.3375         & 0.8009         & 0.6976         & 0.1215         \\
MI-FGSM$_{L_1}$ &
0.5921        *& \rs{0.6541}   *& 0.5543        *& 0.2331        *&
0.3598         & 0.8193         & 0.7130         & 0.1355         &
0.9548         & 0.6024         & 0.1508         & 0.2294         &
0.3266         & 0.8113         & 0.7142         & 0.1180         \\\midrule
TI-MI-FGSM$_\mathrm{\emph{BCE}}$&
0.5380        *& 0.7206        *& 0.5371        *& 0.2112        *&
0.3146         & 0.8197         & 0.7322         & 0.1316         &
0.9623         & 0.5648         & 0.1504         & 0.2318         &
0.3628         & 0.7894         & 0.6784         & 0.1236         \\
TI-MI-FGSM$_{L_1}$&
0.4903        *& 0.7247        *& 0.6074        *& 0.1789        *&
0.3097         & 0.8378         & 0.7514         & 0.1234         &
0.9583         & 0.5677         & 0.1500         & 0.2315         &
0.3489         & 0.8050         & 0.6900         & 0.1190        \\\midrule
ColorFool       &
0.4199         & 0.7962         & 0.6871         & 0.1213         &
0.2169         & 0.8851         & 0.8123         & 0.0903         &
\rt{0.9663}    & 0.5830         & \rt{0.1048}    & 0.2340         &
0.2829         & 0.8366         & 0.7471         & 0.1097         \\\midrule\midrule
Jadena$_\mathrm{\emph{single}}$&
\rt{0.6864}    & 0.7044         & \rs{0.3831}    & 0.1753         &
\rt{0.4868}    & \rt{0.7682}    & \rt{0.6162}    & 0.1529         &
\rt{0.9663}    & \rt{0.5490}    & \rt{0.1199}    & \rt{0.2354}    &
\rt{0.5538}    & \rt{0.6985}    & \rt{0.5152}    & \rt{0.1431}    \\
Jadena$_\mathrm{\emph{group}}$&
0.6849         & 0.7069         & \rt{0.3838}    & 0.1777         &
\rf{0.5221}    & \rs{0.7453}    & \rf{0.5842}    & \rf{0.1619}    &
\rf{0.9826}    & \rs{0.5212}    & \rf{0.0715}    & \rs{0.2392}    &
\rf{0.6060}    & \rs{0.6671}    & \rs{0.4650}    & \rf{0.1545}    \\
Jadena$_\mathrm{\emph{augment}}$&
\rs{0.6893}    & 0.7016         & \rf{0.3747}    & 0.1749         &
\rs{0.5146}    & \rf{0.7432}    & \rs{0.5960}    & \rs{0.1592}    &
\rs{0.9816}    & \rf{0.5137}    & \rs{0.0754}    & \rf{0.2394}    &
\rs{0.6010}    & \rf{0.6648}    & \rf{0.4616}    & \rs{0.1513}    \\
Jedena$_\mathrm{\emph{augment}}$ w/o exposure&
0.5261        & 0.7536        & 0.5483        & 0.1437        &
0.3821        & 0.8185        & 0.7074        & 0.1243        &
0.9533        & 0.6028        & 0.1473        & 0.2290        &
0.4303        & 0.7680        & 0.6347        & 0.1240        \\
\bottomrule
    \end{tabular}
    \end{adjustbox}
    \vspace{-8pt}
    \caption{Comparison with existing attacks. Noise$_{\epsilon}$s apply random additive noise sampled from a uniform distribution $\mathrm{U}\left(-\epsilon,\,\epsilon\right)$ in each channel of the images. The attacks of FGSM and *-FGSMs are performed against co-saliency labels and have full access to the network structure and parameters of the GICD model. *$_{\left\{\mathrm{\emph{BCE}},\,L_1\right\}}$ means that binary cross-entropy (BCE) or $L_1$ loss is adopted as the objective function. We mark white-box attacks with stars and highlight the top three results in \rf{red}, \rs{green}, and \rt{blue} respectively.}
    \label{tab:atk_cmp}\vspace{-15pt}
\end{table*}

\begin{figure*}[tb]
	\centering
	\includegraphics[width=0.8\linewidth]{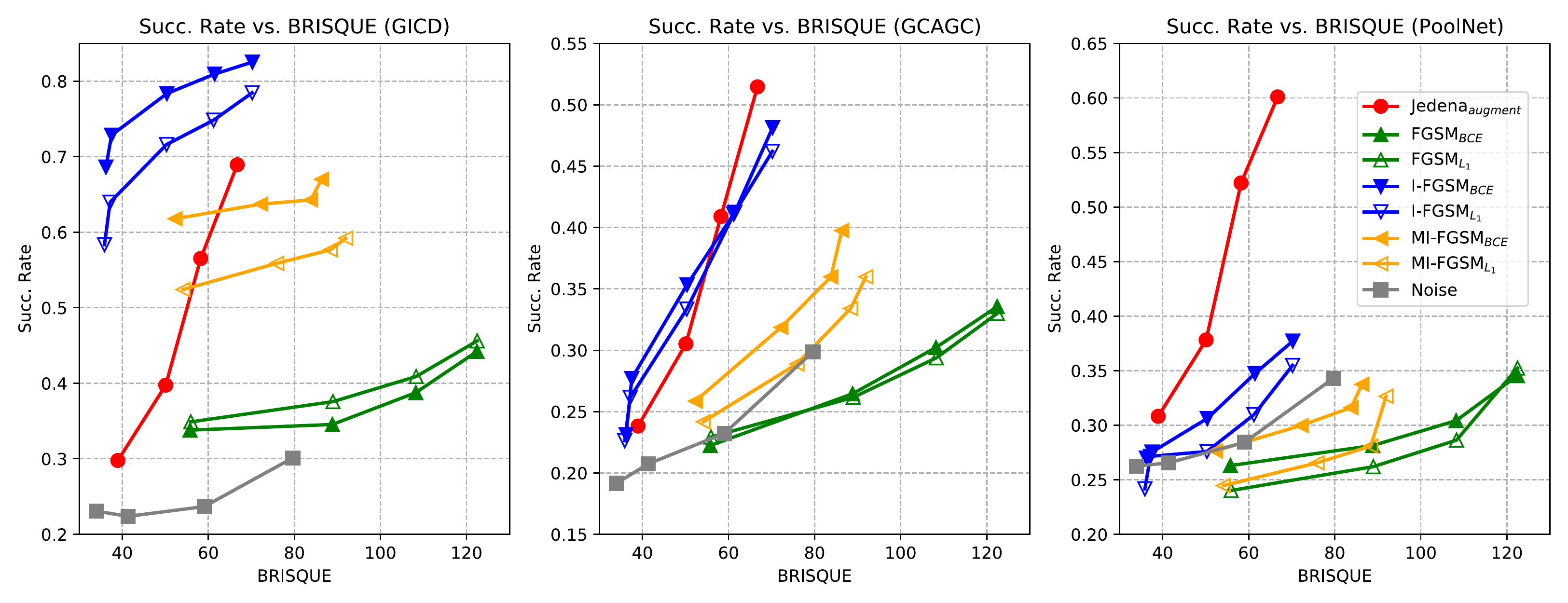}
    \vspace{-8pt}
	\caption{Success rate \vs BRISQUE. This figure shows the change in success rate of attacks with distortion degree. A larger BRISQUE suggests a worse quality of images, \ie, more distortion. Note that, the baselines in the first plot are under white-box settings. The performance of ours is plotted in \textcolor{red}{red}.}
	\label{fig:succ_vs_brisque}\vspace{-8pt}
\end{figure*}

\begin{figure*}[tb]
	\centering
	\includegraphics[width=0.8\linewidth]{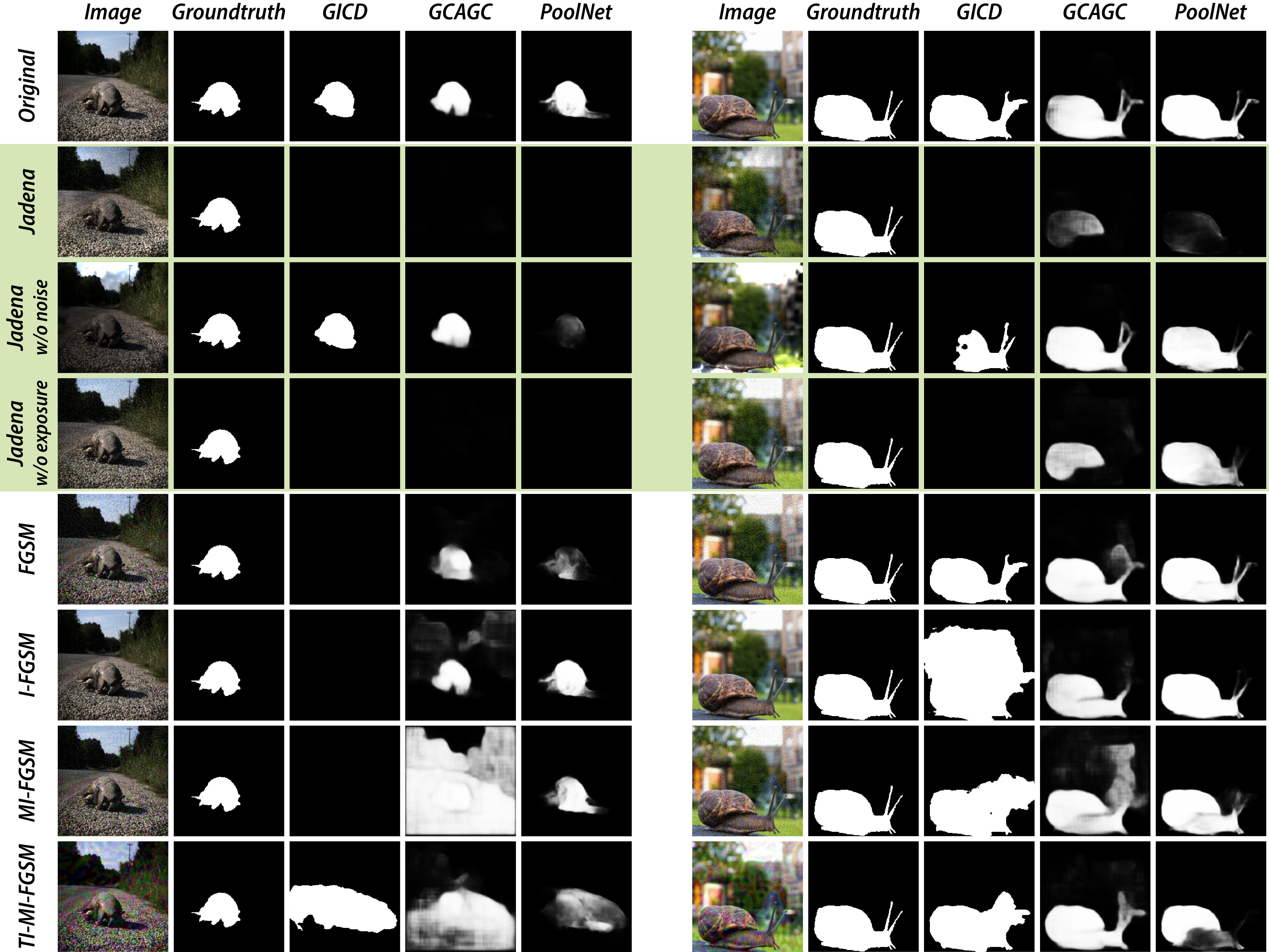}
    \vspace{-7pt}
	\caption{Visualization of attack results. We show two visualized cases in this figure, with the original/perturbed image, the ground-truth of the co-saliency map and the results of GICD, GCAGC, and PoolNet in the columns, and highlight our method in green.}
	\label{fig:atk_cmp_vis}\vspace{-15pt}
\end{figure*}

\subsection{Experimental setups}\label{subsec:exp_set}

{\flushleft\bf Datasets.} We conduct experiments on Cosal2015 \cite{zhang2016detection}, iCoseg \cite{batra2010icoseg}, CoSOD3k \cite{fan2020taking}, and CoCA \cite{zhang2020gradient}. Cosal2015 and CoSOD3k are two large-scale datasets, containing 2,015 and 3,316 images of 50 and 160 groups respectively. iCoseg contains 643 images of 38 groups, in which objects are in the same scene within a group. CoCA is well-designed for evaluating CoSOD methods due to its multiple salient objects in each image. It contains 1,295 images of 80 groups.

{\flushleft\bf Models.} We employ three CoSOD models and one SOD model to evaluate the attacks. GICD \cite{zhang2020gradient} and GCAGC \cite{zhang2020adaptive} are two state-of-the-art CoSOD methods. We employ GCAGC with an HRNet backbone \cite{wang2019hrnet}. PoolNet \cite{liu2019poolnet} is a CNN-based SOD method, but we treat it as a CoSOD method due to its competitive performance on CoSOD datasets. CBCD \cite{fu2013cluster} is a traditional method for both CoSOD and SOD, but we only adopt the co-saliency results.

{\flushleft\bf Metrics.} In our experiments, we employ four metrics to evaluate how adversarial attacks affect detection methods, \ie, average precision (AP) \cite{zhang2018review}, F-measure score $F_\beta$ with $\beta^2=0.3$ \cite{achanta2009frequency}, mean absolute error (MAE) \cite{zhang2018review}, and success rate (S). The first three metrics are widely used for CoSOD evaluations; we can evaluate adversarial attacks by observing how they change from the original performance.
The last metric, success rate, aims to evaluate the ability of an attack to make images undiscoverable. We treat an attack as successful when the IoU between the result of a perturbed image and the corresponding ground-truth map is less than 0.5. IoU is widely used in object detection and visual tracking tasks and defined as
$\mathrm{IoU}=\frac{\text{Aera of Overlap}}{\text{Aera of Union}}$. To calculate success rate, we simply divide the total number of results by the number of successful results. 
Furthermore, to measure the quality of the perturbed images and the perceptibility of the applied perturbations, we employ a no-reference image quality assessor, \ie, BRISQUE \cite{mittal2011brisque}. A smaller BRISQUE score suggests better quality of an image.

\subsection{Attack performance and ablation study}\label{subsec:atk_perf}

We apply our method on several datasets and use four CoSOD methods to predict result maps for the generated adversarial examples. 
We evaluate three variants of the proposed Jadena, with varied reference strategies to build a group around the target image, \ie, no references, references from the same group, and augmented references of original images, denoted by Jadena$_{\{\mathrm{\emph{single}},\,\mathrm{\emph{group}},\,\mathrm{\emph{augment}}\}}$, respectively.
The augmentation of original images includes flipping, mirroring, left rotation, and right rotation, which are designed for full black-box attack, even when images of the same group are not provided.
We optimize Eq.~\eqref{eq:obj_consist} for the first variant and Eq.~\eqref{eq:obj_coconsist} for the other two variants.
Moreover, we apply Jedena$_\mathrm{\emph{augment}}$ without additive noise perturbation or exposure tuning, and report them as Jedena$_\mathrm{\emph{augment}}$ w/o noise and Jedena$_\mathrm{\emph{augment}}$ w/o exposure.

For the feature extraction required for the objective function computation, we employ a general classification model, ResNet50 \cite{he2016resnet}, and adopt the outputs of stage-$\left\{1,2,3\right\}$. For the optimization of our proposed objective function, we use MI-FGSM and adopt $\mu=1.0$ for momentum decay. For other hyperparameters of the proposed method, we set the iteration number $N=20$, $\alpha_n=1$ \wrt pixel values in $[0,\,255]$ and maximum noise-perturbation $\epsilon=16$, following standard additive-perturbation-based attack settings.
For the optimization step size of exposure coefficients and the offset map, we use $\alpha_\mathbf{a}=0.1$ and $\alpha_\mathbf{U}=0.01$.
To smooth the exposure tuning, we use $\lambda_b=0.5$ for the ``single'' variant and $\lambda_b=0.01$ for the ``augment'' or ``group'' variant. $\lambda_s=0.01$ and polynomial degree $D=10$, considering both success rate and imperceptibility of perturbations.
We report the performance across CoSOD methods in Table~\ref{tab:atk_perf}.

We observe that Jedena$_\mathrm{\emph{augment}}$ and Jedena$_\mathrm{\emph{group}}$ achieve the best performance on most metrics.
Even without sampling images from the same group and only augmenting the original images, Jedena$_\mathrm{\emph{augment}}$ is competitive to Jedena$_\mathrm{\emph{group}}$ for GCAGC and PoolNet, and outperforms Jedena$_\mathrm{\emph{group}}$ for GICD. This demonstrates that providing the references is not necessary, as they may be obtained by augmentation.
Jedena$_\mathrm{\emph{augment}}$ performs better than the ``w/o noise'' and ``w/o exposure'' versions, except for CBCD, which suggests that our proposed joint perturbation plays an important role in fooling CoSOD methods.
While it is difficult to handle challenging CoSOD datasets, we find that the traditional method based on color histogram, \ie, CBCD, is much more robust to noise than exposure tuning, since it shows a lower performance for ``w/o noise'' than ``w/o exposure''. In contrast, the other deep methods are more sensitive to noise.

\subsection{Comparison with existing attacks}\label{subsec:atk_cmp}

\begin{figure}[tb]
	\centering
	\includegraphics[width=0.95\linewidth]{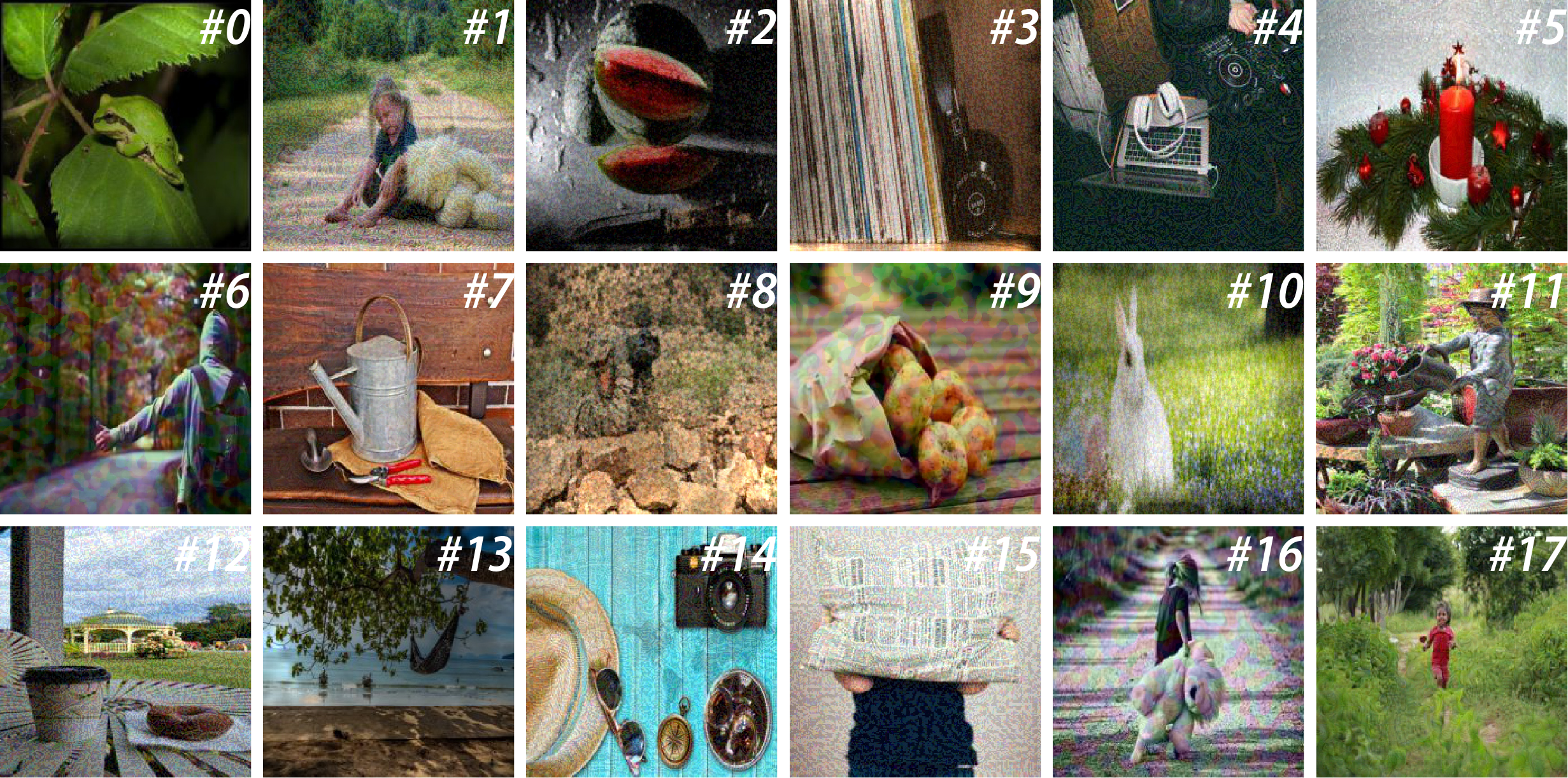}
	\vspace{-7pt}
	\caption{Mixed perturbed images. \#$0$, \#$2$, \#$3$, \#$8$, \#$10$, \#$13$ are generated by the proposed method (\#$0$ and \#$13$ only contain exposure variation) and the others are generated by different baselines (FGSM: \#$1$, \#$12$, \#$15$; I-FGSM: \#$5$, \#$7$, \#$17$; MI-FGSM: \#$4$, \#$11$, \#$14$; TI-MI-FGSM: \#$6$, \#$9$, \#$16$).}
	\label{fig:mixed_perturbed}\vspace{-15pt}
\end{figure}

{\flushleft\bf Baselines.} We consider a series of additive-perturbation-noise attacks as baselines, \ie, FGSM \cite{goodfellow2014explaining}, I-FGSM \cite{kurakin2016adversarial}, MI-FGSM \cite{dong2018boosting}, TI-MI-FGSM \cite{dong2019evading}, and ColorFool \cite{shamsabadi2020colorfool}. For the hyperparameters, we set the max perturbation $\epsilon=16$ with pixel values in $[0,\,255]$, as well as iteration number $N=20$, and step size $\alpha_n=1$ for iteration attacks. Momentum decay $\mu=1.0$ for momentum methods, and the Gaussian kernel length is 15 for TI-MI-FGSM. Since no specific attacks for CoSOD methods have been proposed, we follow the white-box attack pattern and employ $L_1$ norm and binary cross-entropy (BCE) as objective functions.

{\flushleft\bf Attack comparison.} We launch baselines and our method on the Cosal2015 dataset. The hyperparameters of our method follow Sec.~\ref{subsec:atk_perf}. The comparison results are shown in Table~\ref{tab:atk_cmp}.
Our proposed Jedena outperforms all existing additive-perturbation-based attacks, except for white-box attacks, which indicates the better transferability of our method. Since we only borrow several shallow layers from the deep classification model, our proposed method has a lower computational cost than iterative attack methods with a result-based objective function, which unavoidably back-propagate gradients through the whole deep model.
To further clarify the advantages of the proposed attack method over baselines, we can compare our method, \ie, Janena$_\text{augment}$~w/o~exposure that disables the exposure variation, with baseline methods under the same $p$-norm ball with $\epsilon_n=16$. Our method still outperforms the baselines on success rates at most of the time.

{\flushleft\bf Image quality comparison.} We evaluate the image quality of generated adversarial examples and investigate how distortion degree affects the attack performance. For baselines attacks, we follow the default settings in this section. For our method, we adopt the ``augment'' variant, \ie, Jadena$_\mathrm{\emph{augment}}$, which is practical for application in real-world settings. We provide visualizations of how different attack methods perturb an image in Fig.~\ref{fig:atk_cmp_vis} and Fig.~\ref{fig:mixed_perturbed}. Note that, in a real attack scenario, the original images would not be provided. To justify this phenomenon, we randomly mix images perturbed by different attacks in Fig.~\ref{fig:mixed_perturbed}. For each attack method, we also plot success rate \vs BRISQUE in Fig.~\ref{fig:succ_vs_brisque} by changing the maximum noise perturbation $\epsilon_n$ and $\lambda_b$. We do not adopt TI-MI-FGSM since it always applies unnatural coarse-grained noise on images and drastically reduces image quality.
We have three findings: (1) Compared with white-box attacks for GICD, our method has competitive performance with MI-FGSM, however, hold worse performance than I-FGSM. (2) When transferred to GCAGC and PoolNet, our method outperforms most additive-perturbation-based methods. For SOD method PoolNet in particular, our method shows significant advantages over I-FGSM and MI-FGSM. This is reasonable for white-box attacks being adversarial against a CoSOD method instead of an SOD method and thus showing lower transferability. (3) Under the setup in Fig.~\ref{fig:mixed_perturbed}, our adversarial images are imperceptible as the baselines.

\begin{table*}[tb]
    \shrinkrow
    \centering
    \begin{adjustbox}{width=\linewidth,center}
    \begin{tabular}{l|cccc|cccc|cccc|cccc}
    \toprule
    & \multicolumn{4}{c|}{GICD w/ HGD}
    & \multicolumn{4}{c|}{GCAGC w/ HGD}
    & \multicolumn{4}{c|}{CBCD w/ HGD}
    & \multicolumn{4}{c}{PoolNet w/ HGD} \\
    & S $\uparrow$ & AP $\downarrow$ & $F_\beta$ $\downarrow$ & MAE $\uparrow$
    & S $\uparrow$ & AP $\downarrow$ & $F_\beta$ $\downarrow$ & MAE $\uparrow$
    & S $\uparrow$ & AP $\downarrow$ & $F_\beta$ $\downarrow$ & MAE $\uparrow$
    & S $\uparrow$ & AP $\downarrow$ & $F_\beta$ $\downarrow$ & MAE $\uparrow$ \\ \midrule\midrule
Original        &
0.2352         & 0.8595         & 0.7800         & 0.0838         &
0.1881         & 0.8960         & 0.8275         & 0.0814         &
0.9504         & 0.6046         & 0.1530         & 0.2287         &
0.2625         & 0.8449         & 0.7626         & 0.1001         \\\midrule\midrule
Noise$_{8}$     &
0.2645         & 0.8510         & 0.7565         & 0.0905         &
0.1950         & 0.8917         & 0.8234         & 0.0840         &
0.9444         & 0.6068         & 0.1496         & 0.2279         &
0.2566         & 0.8495         & 0.7690         & 0.0992         \\
Noise$_{16}$    &
0.2710         & 0.8464         & 0.7571         & 0.0933         &
0.2283         & 0.8826         & 0.8063         & 0.0913         &
0.9499         & 0.6044         & 0.1512         & 0.2280         &
0.2645         & 0.8458         & 0.7644         & 0.1016         \\
Noise$_{32}$    &
0.3117         & 0.8250         & 0.7238         & 0.1051         &
0.2769         & 0.8585         & 0.7697         & 0.1059         &
0.9548         & 0.6017         & 0.1480         & 0.2292         &
0.3052         & 0.8284         & 0.7335         & 0.1097         \\\midrule\midrule
FGSM$_\mathrm{\emph{BCE}}$&
0.4194        *& 0.7753        *& 0.6439        *& 0.1311        *&
0.3141         & 0.8415         & 0.7476         & 0.1194         &
0.9543         & 0.5953         & 0.1492         & 0.2295         &
0.3136         & 0.8218         & 0.7244         & 0.1133         \\
FGSM$_{L_1}$    &
0.4447        *& 0.7707        *& 0.6205        *& 0.1381        *&
0.3002         & 0.8446         & 0.7570         & 0.1179         &
0.9509         & 0.5978         & 0.1496         & 0.2290         &
0.3176         & 0.8205         & 0.7216         & 0.1131         \\\midrule
I-FGSM$_\mathrm{\emph{BCE}}$&
\rf{0.7012}   *& \rf{0.6639}   *& 0.4141        *& \rf{0.3114}   *&
0.2452         & 0.8583         & 0.7818         & 0.1055         &
0.9459         & 0.6049         & 0.1511         & 0.2282         &
0.2640         & 0.8435         & 0.7639         & 0.1022         \\
I-FGSM$_{L_1}$  &
0.6094        *& \rt{0.6692}   *& 0.5340        *& \rt{0.2519}   *&
0.2357         & 0.8735         & 0.7936         & 0.0988         &
0.9484         & 0.6056         & 0.1544         & 0.2277         &
0.2591         & 0.8451         & 0.7640         & 0.1019         \\\midrule
MI-FGSM$_\mathrm{\emph{BCE}}$ &
0.6491        *& 0.6773        *& 0.4616        *& \rs{0.2749}   *&
0.3687         & 0.8000         & 0.7058         & 0.1424         &
0.9519         & 0.5998         & 0.1479         & 0.2289         &
0.3072         & 0.8154         & 0.7184         & 0.1156         \\
MI-FGSM$_{L_1}$ &
0.5826        *& \rs{0.6649}   *& 0.5562        *& 0.2249        *&
0.3310         & 0.8258         & 0.7363         & 0.1260         &
0.9514         & 0.6012         & 0.1490         & 0.2287         &
0.3037         & 0.8253         & 0.7326         & 0.1130         \\\midrule
TI-MI-FGSM$_\mathrm{\emph{BCE}}$&
0.5444        *& 0.7148        *& 0.5285        *& 0.2066        *&
0.3062         & 0.8162         & 0.7350         & 0.1354         &
0.9593         & 0.5670         & 0.1537         & 0.2310         &
0.3787         & 0.7834         & 0.6666         & 0.1259         \\
TI-MI-FGSM$_{L_1}$&
0.4998        *& 0.7215        *& 0.5987        *& 0.1776        *&
0.2973         & 0.8360         & 0.7594         & 0.1228         &
0.9583         & 0.5680         & 0.1542         & 0.2308         &
0.3628         & 0.8022         & 0.6808         & 0.1207         \\\midrule\midrule
Jadena$_\mathrm{\emph{single}}$&
\rs{0.6988}    & 0.7051         & \rs{0.3721}    & 0.1778         &
\rt{0.4675}    & \rt{0.7630}    & \rt{0.6331}    & \rt{0.1546}    &
\rt{0.9648}    & \rt{0.5500}    & \rt{0.1194}    & \rt{0.2347}    &
\rt{0.4739}    & \rt{0.7335}    & \rt{0.5884}    & \rt{0.1325}    \\
Jadena$_\mathrm{\emph{group}}$&
0.6968         & 0.7106         & \rt{0.3738}    & 0.1797         &
\rs{0.4774}    & \rs{0.7499}    & \rs{0.6205}    & \rf{0.1625}    &
\rf{0.9846}    & \rs{0.5203}    & \rf{0.0664}    & \rs{0.2388}    &
\rf{0.5290}    & \rs{0.7040}    & \rs{0.5273}    & \rf{0.1415}    \\
Jadena$_\mathrm{\emph{augment}}$&
\rt{0.6983}    & 0.7034         & \rf{0.3617}    & 0.1779         &
\rf{0.4893}    & \rf{0.7452}    & \rf{0.6160}    & \rs{0.1611}    &
\rs{0.9831}    & \rf{0.5167}    & \rs{0.0741}    & \rf{0.2390}    &
\rs{0.5266}    & \rf{0.6969}    & \rf{0.5262}    & \rs{0.1399}    \\
\bottomrule
    \end{tabular}
    \end{adjustbox}
    \vspace{-8pt}
    \caption{Defense results with HGD. The settings are the same as Table~\ref{tab:atk_cmp} except the CoSOD methods are equipped with HGD for defense. We mark white-box attacks with stars and highlight top 3 results with \rf{red}, \rs{green}, and \rt{blue} respectively. The attack methods maintains similar performance rankings compared to the attack methods w/o defense which is reported in Table~\ref{tab:atk_cmp}.}
    \label{tab:atk_cmp_def}\vspace{-5pt}
\end{table*}

\begin{table*}[tb]
    \shrinkrow
    \centering
    \begin{adjustbox}{width=\linewidth,center}
    \begin{tabular}{l|cccc|cccc|cccc|cccc}
    \toprule
    & \multicolumn{4}{c|}{PoolNet}
    & \multicolumn{4}{c|}{EGNet}
    & \multicolumn{4}{c|}{BASNet}
    & \multicolumn{4}{c}{U$^2$Net} \\
    & S $\uparrow$ & AP $\downarrow$ & $F_\beta$ $\downarrow$ & MAE $\uparrow$
    & S $\uparrow$ & AP $\downarrow$ & $F_\beta$ $\downarrow$ & MAE $\uparrow$
    & S $\uparrow$ & AP $\downarrow$ & $F_\beta$ $\downarrow$ & MAE $\uparrow$
    & S $\uparrow$ & AP $\downarrow$ & $F_\beta$ $\downarrow$ & MAE $\uparrow$ \\ \midrule\midrule
Original        &
0.1368         & 0.9251         & 0.8546         & 0.0463         &
0.1748         & 0.9246         & 0.8340         & 0.0551         &
0.0809         & 0.9261         & 0.8992         & 0.0369         &
0.0822         & 0.9395         & 0.8928         & 0.0378         \\\midrule\midrule
Noise$_{8}$     &
0.1375         & 0.9263         & 0.8547         & 0.0467         &
0.1831         & 0.9243         & 0.8318         & 0.0555         &
0.0843         & 0.9237         & 0.8983         & 0.0384         &
0.0857         & 0.9386         & 0.8924         & 0.0394         \\
Noise$_{16}$    &
0.1472         & 0.9225         & 0.8469         & 0.0493         &
0.2032         & 0.9218         & 0.8219         & 0.0581         &
0.0940         & 0.9163         & 0.8884         & 0.0418         &
0.0974         & 0.9315         & 0.8861         & 0.0424         \\
Noise$_{32}$    &
0.2163         & 0.8982         & 0.8052         & 0.0616         &
0.2716         & 0.9007         & 0.7793         & 0.0696         &
0.1417         & 0.8912         & 0.8639         & 0.0513         &
0.1700         & 0.9046         & 0.8534         & 0.0560         \\\midrule\midrule
FGSM$_\mathrm{\emph{BCE}}$&
0.5563        *& 0.6948        *& 0.5837        *& 0.1354        *&
0.5308         & 0.7856         & 0.6054         & 0.1154         &
0.3117         & 0.8059         & 0.7743         & 0.0841         &
0.2999         & 0.8320         & 0.7637         & 0.0881         \\
FGSM$_{L_1}$    &
0.2191        *& 0.8556        *& 0.7740        *& 0.0733        *&
0.2716         & 0.8773         & 0.7701         & 0.0746         &
0.1679         & 0.8751         & 0.8459         & 0.0565         &
0.1638         & 0.8887         & 0.8402         & 0.0594         \\\midrule
I-FGSM$_\mathrm{\emph{BCE}}$&
0.1576        *& 0.9158        *& 0.8392        *& 0.0518        *&
0.2232         & 0.9104         & 0.8023         & 0.0631         &
0.1196         & 0.8986         & 0.8654         & 0.0502         &
0.1237         & 0.9146         & 0.8644         & 0.0505         \\
I-FGSM$_{L_1}$  &
0.1548        *& 0.9202        *& 0.8406        *& 0.0499        *&
0.2101         & 0.9153         & 0.8088         & 0.0609         &
0.1140         & 0.9073         & 0.8726         & 0.0471         &
0.1099         & 0.9228         & 0.8720         & 0.0477         \\\midrule
MI-FGSM$_\mathrm{\emph{BCE}}$&
\rf{0.9364}   *& \rf{0.3619}   *& \rf{0.3078}   *& \rf{0.4503}   *&
\rt{0.5764}    & 0.7799         & 0.5688         & \rt{0.1169}    &
0.3172         & 0.8125         & 0.7815         & 0.0822         &
0.3338         & 0.8249         & 0.7593         & 0.0899         \\
MI-FGSM$_{L_1}$ &
\rt{0.7146}   *& \rt{0.6320}   *& 0.4873        *& \rt{0.2570}   *&
0.4340         & 0.8077         & 0.6556         & 0.1070         &
0.2668         & 0.8059         & \rt{0.7718}    & \rt{0.0871}    &
0.2875         & 0.8266         & 0.7582         & 0.0923         \\\midrule
TI-MI-FGSM$_\mathrm{\emph{BCE}}$&
\rs{0.8466}   *& \rs{0.4869}   *& \rs{0.3979}   *& \rs{0.3467}   *&
\rs{0.6683}    & \rs{0.7432}    & \rs{0.4952}    & \rs{0.1310}    &
\rt{0.3877}    & \rt{0.7761}    & \rs{0.7530}    & \rs{0.0922}    &
\rt{0.3704}    & \rt{0.8107}    & \rt{0.7480}    & \rs{0.0940}    \\
TI-MI-FGSM$_{L_1}$&
0.5197        *& 0.7156        *& 0.5794        *& 0.1790        *&
0.4202         & 0.8047         & 0.6550         & 0.1089         &
0.2605         & 0.8190         & 0.7747         & 0.0820         &
0.2370         & 0.8474         & 0.7777         & 0.0827         \\\midrule\midrule
Jadena$_\mathrm{\emph{single}}$&
0.5771         & 0.7287         & 0.5179         & 0.1136         &
0.5722         & \rt{0.7706}    & \rt{0.5337}    & 0.1123         &
\rs{0.3891}    & \rs{0.7666}    & 0.7837         & 0.0866         &
\rs{0.4181}    & \rs{0.7948}    & \rs{0.7405}    & \rt{0.0932}    \\
Jadena$_\mathrm{\emph{augment}}$&
0.6745         & 0.6645         & \rt{0.4185}    & 0.1298         &
\rf{0.7091}    & \rf{0.7062}    & \rf{0.3997}    & \rf{0.1325}    &
\rf{0.4983}    & \rf{0.7144}    & \rf{0.7412}    & \rf{0.1030}    &
\rf{0.5121}    & \rf{0.7565}    & \rf{0.7006}    & \rf{0.1061}    \\
\bottomrule
    \end{tabular}
    \end{adjustbox}
    \vspace{-8pt}
    \caption{Comparison with existing attacks against SOD. Noise$_{\epsilon}$s apply random additive noise sampled from uniform distribution $\mathrm{U}\left(-\epsilon,\,\epsilon\right)$ on each channel. The attacks of FGSM and *-FGSMs are launched against single saliency labels in a white-box manner. *$_{\left\{\mathrm{\emph{BCE}},\,L_1\right\}}$ means that binary cross-entropy (BCE) or $L_1$ loss is adopted as the objective function. We mark white-box attacks with stars and highlight top 3 results with \rf{red}, \rs{green}, and \rt{blue} respectively.}
    \label{tab:atk_cmp_sod}\vspace{-5pt}
\end{table*}

{\flushleft\bf Defense.} We adopt a defense method, \ie, HGD \cite{liao2018defense}, to further demonstrate the effectiveness of the proposed method and report the results in Table~\ref{tab:atk_cmp_def}. We find that the proposed method still outperforms the baselines except for the white-box settings. We report more defense results with other defense methods in the supplementary materials.

{\flushleft\bf Attack comparison against SOD methods.} As shown in Table~\ref{tab:atk_cmp_sod}, we also apply the proposed method against SOD methods and report the performance on the DUTS-TE \cite{wang2017duts} dataset. We employ white-box attacks against PoolNet \cite{liu2019poolnet} as the baseline and evaluate all attacks with PoolNet, EGNet \cite{zhao2019EGNet}, BASNet \cite{qin2019BASNet}, and U$^2$Net \cite{qin2019U2Net}. Our proposed methods achieve competitive performance and higher transferability across SOD methods than the white-box attacks. It indicates that our proposed loss function and joint perturbation can be adopted to attack SOD methods.

\section{Conclusion}\label{sec:concl}

In this work, we have investigated a novel problem: how to effectively protect sensitive image content from being discovered by the state-of-the-art CoSOD methods. We have addressed this problem from the perspective of adversarial attacks and identified a novel task, \ie, adversarial co-saliency attack. Given an image selected from a group of images containing some common and salient objects, the aim of this task is to generate an adversarial version of the image that can mislead CoSOD methods to predict incorrect co-salient regions, thus evading co-saliency detection and making the images undiscoverable by the co-saliency detection algorithms. We have proposed the very first black-box \emph{joint adversarial exposure and noise attack (Jadena)}, where we jointly and locally tune the exposure and additive perturbations of the image according to a newly designed \emph{high-feature-level contrast-sensitive loss} function. Our method, without needing any information on the state-of-the-art CoSOD methods, leads to significant performance degradation on various co-saliency detection datasets and makes the co-salient objects undetectable. The effectiveness of the proposed method is thoroughly validated and showcased through various quantitative and qualitative experiments. We believe that the proposed method can greatly benefit the design and execution of privacy-oriented content sharing in the public domain where malicious extractions can be effectively inhibited. 

\vspace{1em}
\noindent\textbf{Acknowledgment:} This work was supported by the National Key Research and Development Program of China (No.~2020YFC1522700), the National Natural Science Foundation of China (No.~62072334), and the General Project of Tianjin (No.~18JCYBJC15200). 
It was also supported in part by the National Research Foundation, Singapore under its AI Singapore Programme (AISG Award No: AISG2-RP-2020-019), Singapore National Cybersecurity R\&D Program No. NRF2018NCR-NCR005-0001, National Satellite of Excellence in Trustworthy Software System No.~NRF2018NCR-NSOE003-0001, NRF Investigatorship No.~NRFI06-2020-0022-0001, and A*STAR AI3 HTPO Seed Fund (C211118012).
We gratefully acknowledge the support of NVIDIA AI Tech Center (NVAITC) and AWS Cloud Credits for Research Award.




\setcounter{figure}{0}
\setcounter{table}{0}
\renewcommand{\thefigure}{\Roman{figure}}
\renewcommand{\thetable}{\Roman{table}}

\appendix
\section{Appendix}\label{sec:appendix}

In this material, we add
a new defense method (\ie, ZeroDCE\cite{guo2020zerodce}). 
The ZeroDCE is able to enhance the low-light images and can be regarded as a defense method against our exposure attack.
Then, we report more experimental results of our method and baselines against four CoSOD methods with the attack, defense, and visualization results. 
Both these demonstrate the advantages of our method over the baseline attacks.

\subsection{More Defense Results}

We further adopt the image enhancement method (\ie, ZeroDCE \cite{guo2020zerodce}) as a defense method and report the results on the Cosal2015 dataset \cite{zhang2016detection} in Table~\ref{tab:atk_cmp_def_more}. The attack methods maintain similar performance rankings compared to the attack methods w/o defense which is reported in Table~\ref{tab:atk_cmp}, demonstrating that the proposed attack is more difficult to be defended than baseline attacks and should be further studied in the future. 

\begin{table*}[htb]
    \centering
    \begin{adjustbox}{width=\linewidth,center}
    \begin{tabular}{l|cccc|cccc|cccc|cccc}
    \toprule
    & \multicolumn{4}{c|}{GICD w/ ZeroDCE}
    & \multicolumn{4}{c|}{GCAGC w/ ZeroDCE}
    & \multicolumn{4}{c|}{CBCD w/ ZeroDCE}
    & \multicolumn{4}{c}{PoolNet w/ ZeroDCE} \\
    & S $\uparrow$ & AP $\downarrow$ & $F_\beta$ $\downarrow$ & MAE $\uparrow$
    & S $\uparrow$ & AP $\downarrow$ & $F_\beta$ $\downarrow$ & MAE $\uparrow$
    & S $\uparrow$ & AP $\downarrow$ & $F_\beta$ $\downarrow$ & MAE $\uparrow$
    & S $\uparrow$ & AP $\downarrow$ & $F_\beta$ $\downarrow$ & MAE $\uparrow$ \\ \midrule\midrule
Original        &
0.2352         & 0.8595         & 0.7800         & 0.0838         &
0.1881         & 0.8960         & 0.8275         & 0.0814         &
0.9504         & 0.6046         & 0.1530         & 0.2287         &
0.2625         & 0.8449         & 0.7626         & 0.1001         \\\midrule\midrule
Noise$_{8}$     &
0.2859         & 0.8356         & 0.7379         & 0.0996         &
0.2342         & 0.8841         & 0.8073         & 0.0926         &
0.9687         & 0.6049         & 0.1058         & 0.2312         &
0.2556         & 0.8472         & 0.7653         & 0.1007         \\
Noise$_{16}$    &
0.3072         & 0.8267         & 0.7281         & 0.1038         &
0.2670         & 0.8695         & 0.7836         & 0.1021         &
0.9742         & 0.5974         & 0.0912         & 0.2322         &
0.2705         & 0.8399         & 0.7599         & 0.1050         \\
Noise$_{32}$    &
0.3926         & 0.7912         & 0.6631         & 0.1214         &
0.3429         & 0.8284         & 0.7337         & 0.1311         &
\rs{0.9836}    & 0.5798         & \rf{0.0743}    & 0.2351         &
0.3206         & 0.8178         & 0.7179         & 0.1145         \\\midrule\midrule
FGSM$_\mathrm{\emph{BCE}}$&
0.5444        *& 0.7323        *& 0.5452        *& 0.1567        *&
0.3816         & 0.8142         & 0.7102         & 0.1418         &
0.9777         & 0.5770         & 0.0875         & 0.2344         &
0.3390         & 0.8091         & 0.7091         & 0.1161         \\
FGSM$_{L_1}$    &
0.5653        *& 0.7310        *& 0.5242        *& 0.1624        *&
0.3891         & 0.8183         & 0.7105         & 0.1429         &
0.9777         & 0.5796         & 0.0911         & 0.2342         &
0.3454         & 0.8068         & 0.7021         & 0.1173         \\\midrule
I-FGSM$_\mathrm{\emph{BCE}}$&
0.7241        *& \rf{0.6492}   *& 0.4056        *& \rf{0.3146}   *&
0.2998         & 0.8415         & 0.7565         & 0.1164         &
0.9737         & 0.6022         & 0.1052         & 0.2312         &
0.2690         & 0.8367         & 0.7552         & 0.1061         \\
I-FGSM$_{L_1}$  &
0.6278        *& \rt{0.6602}   *& 0.5221        *& \rt{0.2551}   *&
0.2918         & 0.8567         & 0.7641         & 0.1131         &
0.9707         & 0.6046         & 0.1059         & 0.2312         &
0.2630         & 0.8406         & 0.7616         & 0.1051         \\\midrule
MI-FGSM$_\mathrm{\emph{BCE}}$ &
0.6660        *& 0.6675        *& 0.4521        *& \rs{0.2629}   *&
0.4323         & 0.7812         & 0.6654         & \rt{0.1627}    &
0.9762         & 0.5860         & 0.0962         & 0.2330         &
0.3409         & 0.7982         & 0.7025         & 0.1215         \\
MI-FGSM$_{L_1}$ &
0.6065        *& \rs{0.6509}   *& 0.5462        *& 0.2192        *&
0.4144         & 0.8033         & 0.6894         & 0.1466         &
0.9762         & 0.5862         & 0.0918         & 0.2330         &
0.3236         & 0.8087         & 0.7141         & 0.1190         \\\midrule
TI-MI-FGSM$_\mathrm{\emph{BCE}}$&
0.5821        *& 0.6968        *& 0.5150        *& 0.1899        *&
0.3672         & 0.8015         & 0.7067         & 0.1440         &
0.9797         & 0.5356         & 0.0886         & 0.2355         &
0.3970         & 0.7721         & 0.6488         & 0.1295         \\
TI-MI-FGSM$_{L_1}$&
0.5538        *& 0.7099        *& 0.5569        *& 0.1734        *&
0.3598         & 0.8196         & 0.7244         & 0.1361         &
0.9826         & \rt{0.5346}    & \rt{0.0778}    & \rt{0.2360}    &
0.3866         & 0.7864         & 0.6651         & 0.1248         \\\midrule\midrule
Jadena$_\mathrm{\emph{single}}$&
\rt{0.7702}    & 0.6793         & \rt{0.3046}    & 0.1903         &
\rt{0.5107}    & \rt{0.7589}    & \rt{0.6101}    & 0.1579         &
0.9792         & 0.5508         & 0.0930         & 0.2355         &
\rt{0.4958}    & \rt{0.7275}    & \rt{0.5755}    & \rt{0.1372}    \\
Jadena$_\mathrm{\emph{group}}$&
\rs{0.7811}    & 0.6776         & \rs{0.2987}    & 0.1951         &
\rs{0.5365}    & \rs{0.7416}    & \rf{0.5750}    & \rf{0.1659}    &
\rt{0.9836}    & \rs{0.5285}    & \rs{0.0771}    & \rs{0.2367}    &
\rs{0.5320}    & \rs{0.7135}    & \rf{0.5342}    & \rf{0.1441}    \\
Jadena$_\mathrm{\emph{augment}}$&
\rf{0.7921}    & 0.6715         & \rf{0.2890}    & 0.1938         &
\rf{0.5439}    & \rf{0.7334}    & \rs{0.5893}    & \rs{0.1645}    &
\rf{0.9841}    & \rf{0.5236}    & 0.0784         & \rf{0.2373}    &
\rf{0.5370}    & \rf{0.7043}    & \rs{0.5347}    & \rs{0.1437}    \\
\bottomrule
    \end{tabular}
    \end{adjustbox}
    \caption{Defense results with ZeroDCE. Noise$_{\epsilon}$s apply random additive noise sampled from a uniform distribution $\mathrm{U}\left(-\epsilon,\,\epsilon\right)$ in each channel of the images. The attacks of FGSM and *-FGSMs are performed against co-saliency labels and have full access to the network structure and parameters of the GICD model. *$_{\left\{\mathrm{\emph{BCE}},\,L_1\right\}}$ means that binary cross-entropy (BCE) or $L_1$ loss is adopted as the objective function. We mark white-box attacks with stars and highlight the top three results in \rf{red}, \rs{green}, and \rt{blue} respectively.}
    \label{tab:atk_cmp_def_more}
\end{table*}

\subsection{More Visualization Results}

\begin{figure*}[htb]
	\centering
	\includegraphics[width=\linewidth]{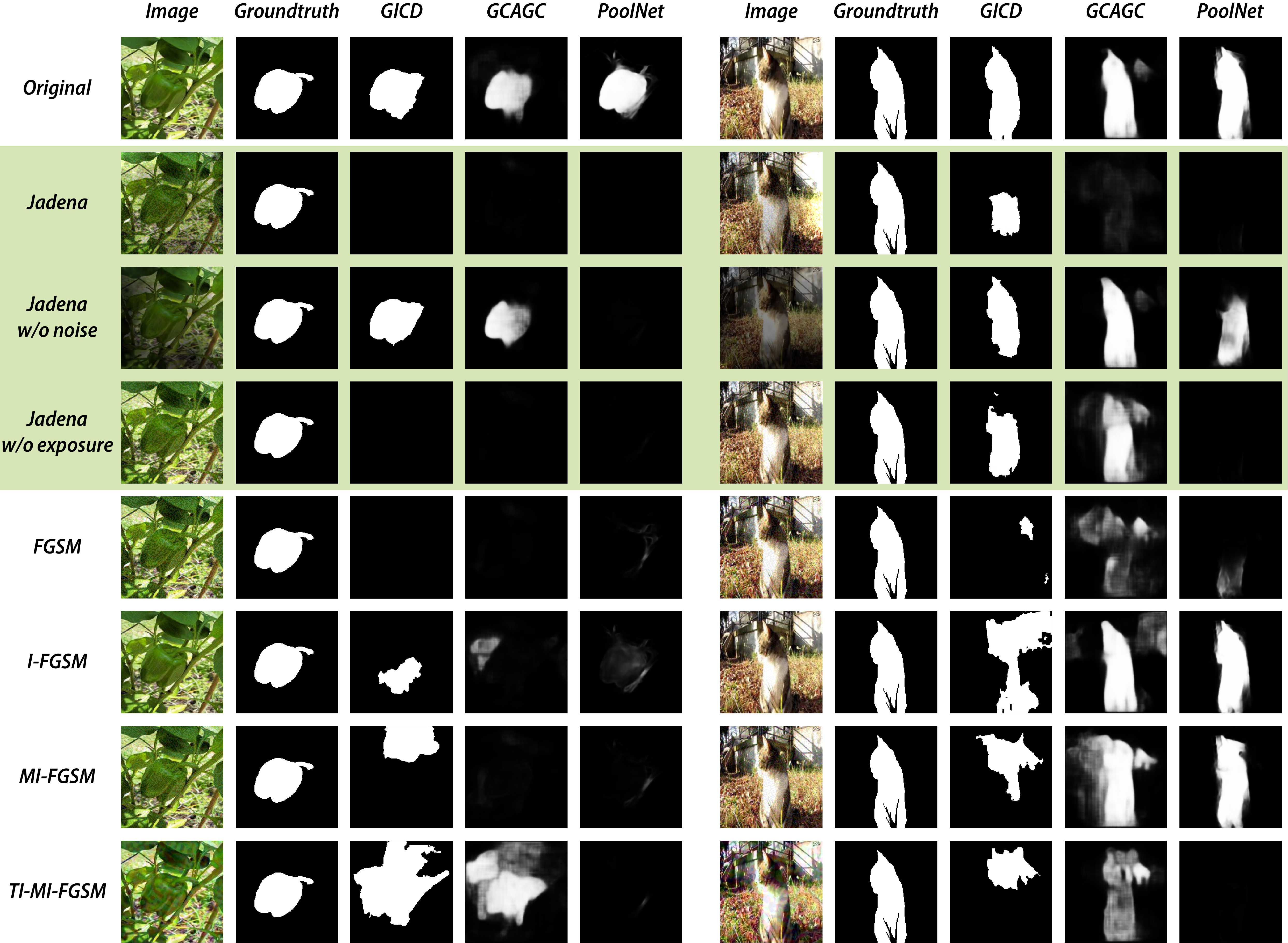}
	\caption{More visualization results from Cosal2015 and CoSOD3k. We visualize the image, the ground-truth of co-saliency map and results of GICD, GCAGC and PoolNet along columns, and show the original images and images perturbed by Jadena, baselines followed by corresponding results for each row. We highlight our method in green.}
	\label{fig:atk_comp_vis_1}
	\vspace{-10pt}
\end{figure*}

\begin{figure*}[htb]
	\centering
	\includegraphics[width=\linewidth]{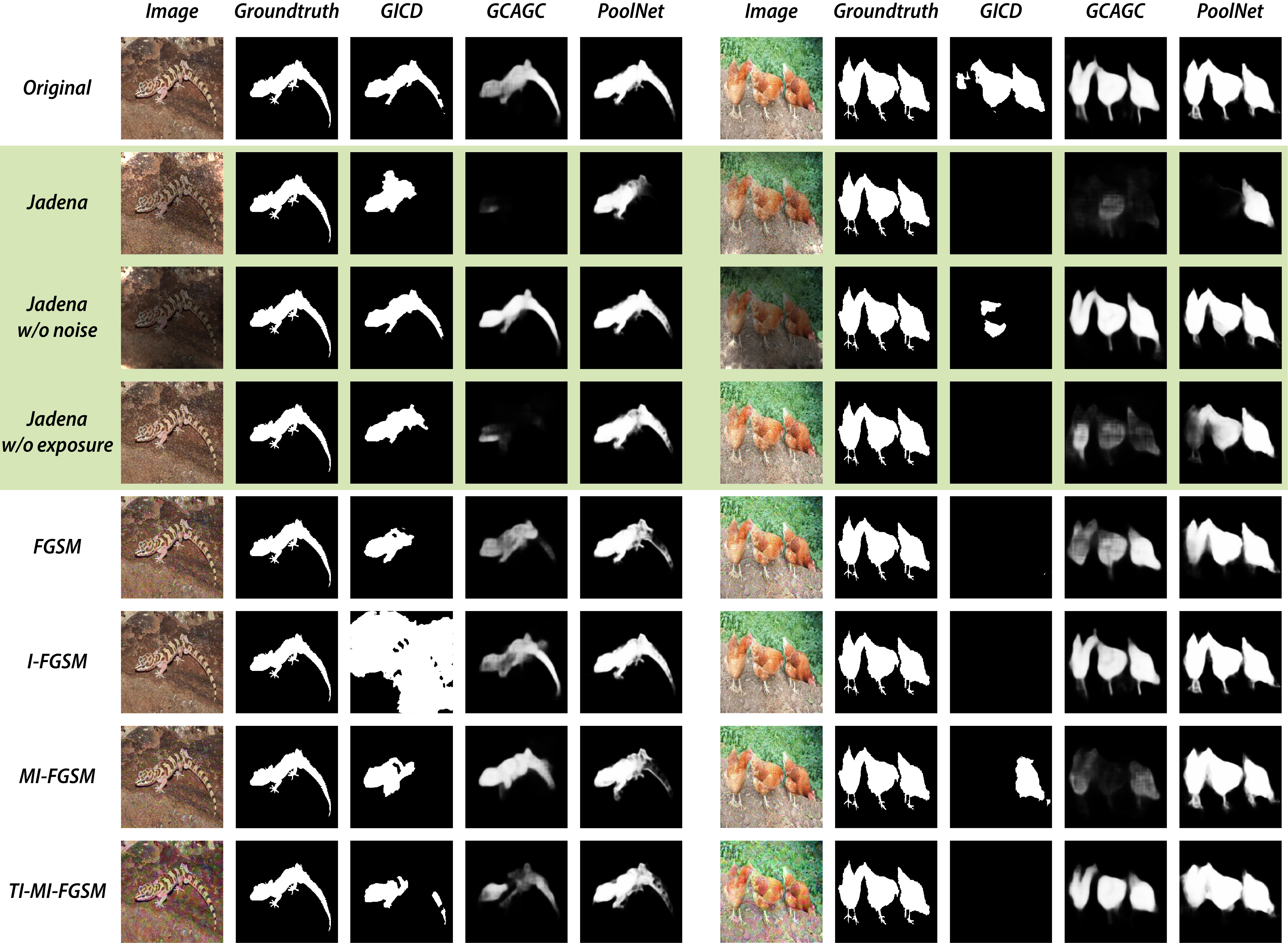}
	\caption{More visualization results from Cosal2015 and CoSOD3k. We visualize the image, the ground-truth of co-saliency map and results of GICD, GCAGC and PoolNet along columns, and show the original images and images perturbed by Jadena, baselines followed by corresponding results for each row. We highlight our method in green.}
	\label{fig:atk_comp_vis_2}
	\vspace{-10pt}
\end{figure*}

\begin{figure*}[htb]
	\centering
	\includegraphics[width=\linewidth]{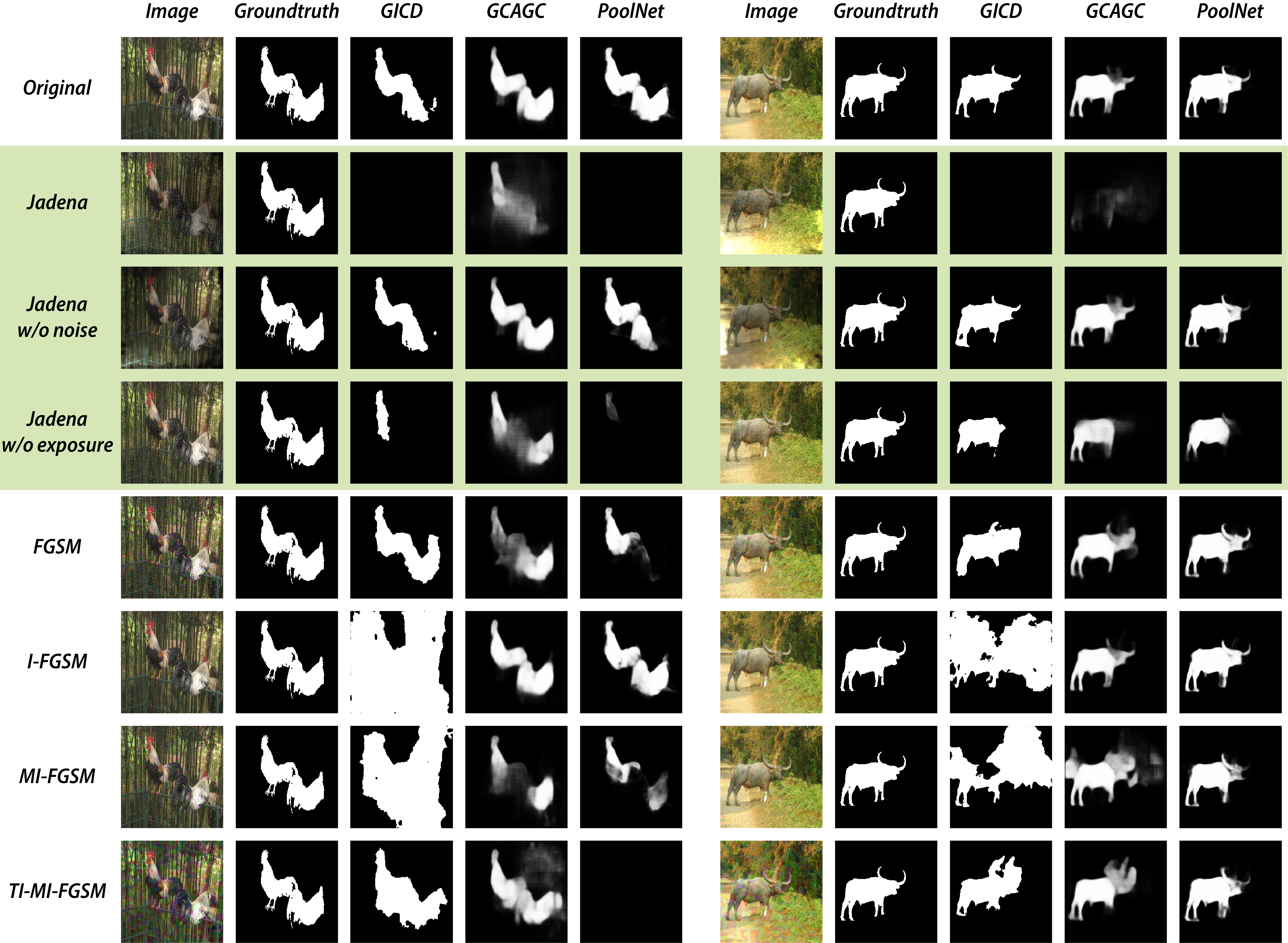}
	\caption{More visualization results from Cosal2015 and CoSOD3k. We visualize the image, the ground-truth of co-saliency map and results of GICD, GCAGC and PoolNet along columns, and show the original images and images perturbed by Jadena, baselines followed by corresponding results for each row. We highlight our method in green.}
	\label{fig:atk_comp_vis_3}
	\vspace{-10pt}
\end{figure*}

\begin{figure*}[htb]
	\centering
	\includegraphics[width=\linewidth]{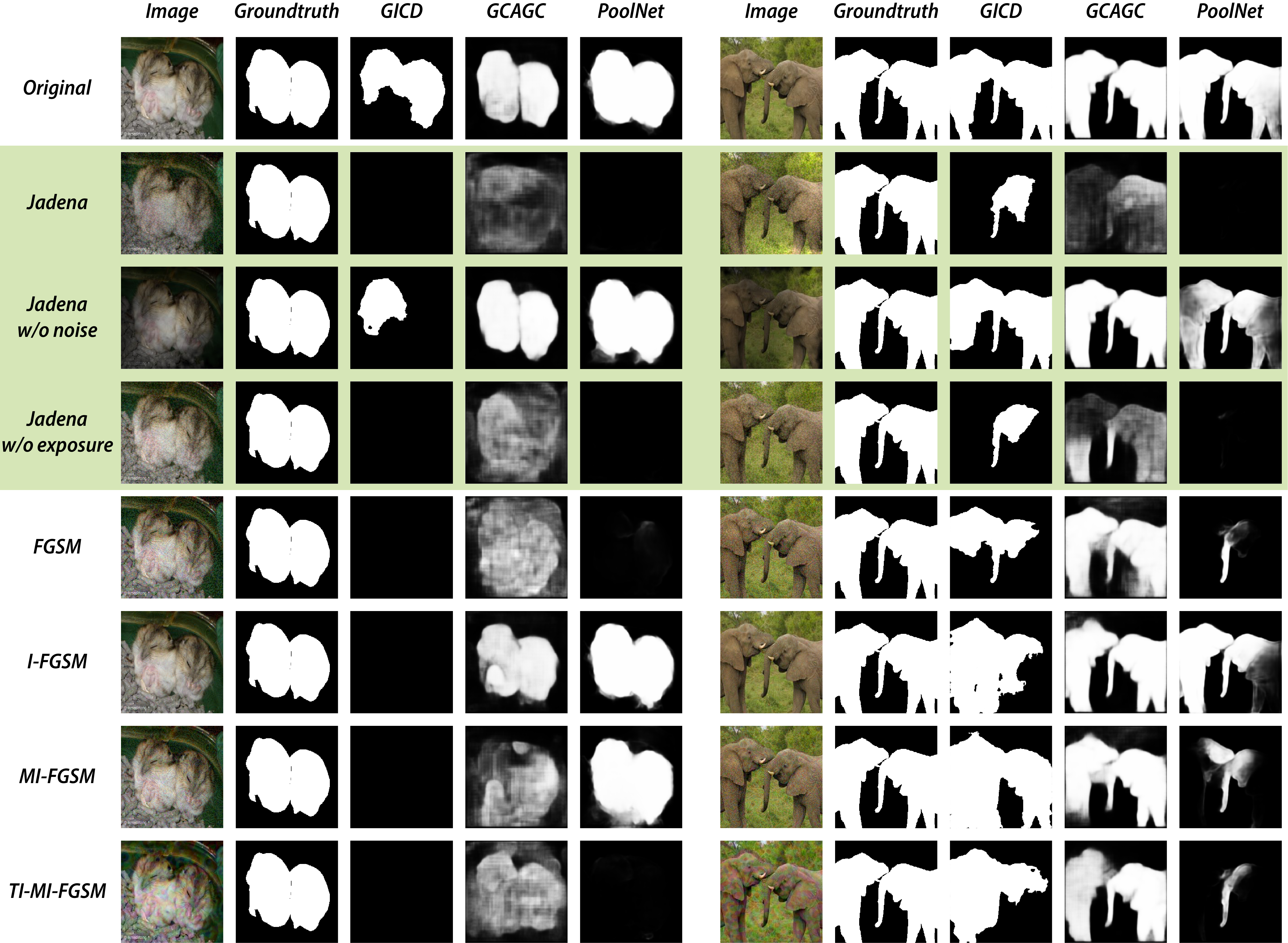}
	\caption{More visualization results from Cosal2015 and CoSOD3k. We visualize the image, the ground-truth of co-saliency map and results of GICD, GCAGC and PoolNet along columns, and show the original images and images perturbed by Jadena, baselines followed by corresponding results for each row. We highlight our method in green.}
	\label{fig:atk_comp_vis_4}
	\vspace{-10pt}
\end{figure*}

We show more visualization results of cases from Cosal2015 \cite{zhang2016detection} and CoSOD3k \cite{fan2020taking} in Figs.~\ref{fig:atk_comp_vis_1}, \ref{fig:atk_comp_vis_2}, ~\ref{fig:atk_comp_vis_3} and \ref{fig:atk_comp_vis_4}. The hyperparameter settings and the layout keep the same with Fig.~\ref{fig:atk_cmp_vis} of the main paper. We may observe adversarial examples generated by our proposed method hold more transferability across CoSOD methods, without perceptible perturbation. Moreover, the visualization results validate the joint perturbation plays an important role in fooling CoSOD methods since our method with joint perturbation outperforms ``w/o noise'' and ``w/o exposure'' versions. We can also notice that TI-MI-FGSM applies unnatural noise, which is more perceptible than any other noise-based attacks.

{\small
\bibliographystyle{ieee_fullname}
\bibliography{ref}
}

\end{document}